\documentclass[letterpaper,journal]{IEEEtran}
\usepackage{setspace}
\usepackage{lineno}

\ifCLASSINFOpdf
\else
\fi

\usepackage{lineno,hyperref}
\usepackage{amsmath,amssymb}
\usepackage{xcolor}
\usepackage{graphicx}
\usepackage[ruled,vlined]{algorithm2e}
\usepackage{algorithmic}
\usepackage[utf8]{inputenc}
\usepackage{longtable}
\usepackage{supertabular,booktabs}
\DeclareMathOperator*{\argmax}{arg\,max}
\usepackage{mathtools}

\usepackage{cancel}
\usepackage[normalem]{ulem}

\begin{document}
%
\title{Optimal Path Planning of Autonomous Marine Vehicles in Stochastic Dynamic Ocean Flows using a GPU-Accelerated Algorithm}
%
%
%

\author{Rohit~Chowdhury,
        Deepak~N.~Subramani,~\IEEEmembership{Member,~IEEE}
\thanks{R. Chowdhury and D. N. Subramani are with the Department
of Computational and Data Sciences, Indian Institute of Science Bangalore, India, 560012 e-mail: deepakns@iisc.ac.in.}
\thanks{This is the first revised version of the manuscript under review in IEEE Journal of Oceanic Engineering}}

%
%

\markboth{IEEE Journal of Oceanic Engineering}%
{Chowdhury and Subramani: GPU-accelerated Optimal Path Planning}
%




\maketitle

\begin{abstract}
Autonomous marine vehicles play an essential role in many ocean science and engineering applications. 
Planning time and energy optimal paths for these vehicles to navigate in stochastic dynamic ocean environments is essential to reduce operational costs. In some missions, they must also harvest solar, wind, or wave energy (modeled as a stochastic scalar field) and move in optimal paths that minimize net energy consumption. Markov Decision Processes (MDPs) provide a natural framework for sequential decision making for robotic agents in such environments. However, building a realistic model and solving the modeled MDP
becomes computationally expensive in large-scale real-time applications, warranting the need of parallel algorithms and efficient implementation. In the present work, we introduce an efficient end-to-end GPU-accelerated algorithm that \textit{(i)} builds the MDP model (computing transition probabilities and expected one-step rewards);
and \textit{(ii)} solves the MDP to compute an optimal policy. We develop methodical and algorithmic solutions to overcome the limited global memory of GPUs by \textit{(i)} using a dynamic reduced-order representation 
of the ocean flows, \textit{(ii)} leveraging the sparse nature of the state transition probability matrix, \textit{(iii)} introducing a neighbouring sub-grid concept and
\textit{(iv)} proving that it is sufficient to use only the stochastic scalar field's mean to compute the expected one-step rewards for missions involving energy harvesting from the environment; thereby saving memory and reducing the computational effort. 
We demonstrate the algorithm on a simulated stochastic dynamic environment and highlight that it builds the MDP model and computes the optimal policy 600-1000x faster than conventional CPU implementations,
making it suitable for real-time use. 
\end{abstract}


%
\IEEEpeerreviewmaketitle

\section{Introduction}
%
%
%
%
The advancement of ocean science and engineering relies heavily on accessing challenging ocean regions using autonomous or unmanned underwater vehicles. These agents perform complex and long-duration missions \cite{bellingham_rajan_2007}, including scientific data collection, defense and security, ocean conservation, and ship-hull damage inspection. A pressing need for such tasks is to operate the autonomous agents optimally such that operational costs are minimized and utility is maximized. An optimal plan (or policy) is one that minimizes a cost function, typically, the expected mission time or energy consumption, while ensuring the safety of the agent. Additionally, the net energy consumption must be minimized for certain vehicles that continuously recharge their batteries by collecting solar, wind or wave energy during a mission. A key challenge in planning optimal paths of these agents is that they get advected by the strong, dynamic and uncertain ocean currents. As such, accurate modeling of the environment and an efficient decision-making framework under uncertainty for the autonomous agent is crucial for successful planning. The planner must also be implemented using fast and scalable algorithms so that it may be utilized in real-time. 

In the present paper, we consider an optimal high-level path planning problem for an autonomous marine agent that has to travel from a given start point to arrive at a target by moving through a stochastic and dynamic ocean environment. We are interested in three different single-objective missions, viz., minimizing the expected mission time, expected energy consumption without environmental harvesting and expected net energy consumption with environmental harvesting. The environment consists of \textit{(i)} a stochastic, dynamic velocity field representing the ocean currents that advect the autonomous agent, \textit{(ii)} a stochastic, dynamic scalar field representing a quantity that must be collected by the agent (e.g., solar radiation or wind energy), and \textit{(iii)} dynamic obstacles that must be avoided. Here, we rely on a reliable prediction model for the environment with quantified uncertainty, which is a reasonably good assumption for underwater applications \cite{lermusiaux_et_al_springer_HOE2016,nemo_book,shchepetkin2005regional,subramani_PhDThesis2018}. 
Further, we focus on developing a computationally efficient algorithm to compute an optimal high-level policy in an offline setting (open-loop control). Our purpose is to use this algorithm in the future to develop an on-board routing scheme where the optimal policy is periodically updated through in-mission environment observations (closed-loop control).

In the above setting, the path planning problem is posed as a finite horizon, undiscounted, total cost Markov Decision Process (MDP) over a discretized spatio-temporal state space (Sec.~\ref{sec:MDPFramework}). 
Here, the uncertain velocity field directly affects the agent’s motion, the uncertain scalar field affects the reward structure that incentivizes the agent’s behavior, and the agent must avoid any obstacle in the mission. In the MDP framework, the objective is to compute an optimal policy that maximizes the expected sum of rewards.
A dynamic programming algorithm may be used to provide an exact solution to the optimization problem. However, for large problems sizes, solving an MDP through dynamic programming becomes computationally expensive \cite{bertsekas1995dynamic}. For a finite horizon MDP, the computational cost of the value iteration algorithm (a dynamic programming algorithm) is $O(n|\mathcal{A}||\mathcal{S}|^2)$, where $n$, $|\mathcal{A}|$, $|\mathcal{S}|$ are the time horizon, action space size and state space size, respectively. Moreover, the computational cost of building the transition probability from ensembles of the stochastic environment forecast is $O(c|N_{r,v}||\mathcal{A}||\mathcal{S}|)$, where $c$ is a constant and $|N_{r,v}|$ is the number of discrete environmental flow ensembles. This cost limits the application of such exact methods for real-time use with large scale problems. However, if fast, scalable algorithms to build MDPs and solve them are available, it opens up possibilities of their use in high-level path planning and also for online re-planning when combined with a data assimilation scheme. To overcome the limitation posed by computational time requirement, we propose an end-to-end Graphical Processing Unit (GPU)-accelerated parallel algorithm to build an MDP and compute an optimal policy in a fraction of time required for serial algorithms on a CPU. The GPU’s parallel processing capability is ideal for this task as several computations in building the MDP transition probabilities and the dynamic programming solution algorithm are embarrassingly parallel. 

The novel contributions of this paper are: \textit{(1)} Development and implementation of a computationally efficient end-to-end GPU-based algorithm to solve the optimal path planning problem using an MDP framework. We highlight the algorithmic challenges that arise in the GPU implementation for building the MDP model and how these challenges are overcome with new optimized Compute Unified Device Architecture (CUDA) kernels; \textit{(2)} Introduction of an additional stochastic scalar field that the agent tries to collect as it moves around during mission. This field affects the reward structure of the autonomous agent, thereby introducing an additional layer of uncertainty. 
In doing so, we have developed an interdisciplinary solution that combines the rigorous MDP framework for planning under uncertainty with a stochastic environment modelling technique and high performance computing.

The organization of the present paper is as follows. First, we briefly discuss the previous work in the field. Next, we describe the theory of Markov Decision Processes (Sec.~\ref{sec:MDPFramework}) and stochastic environmental modeling (Sec.~\ref{sec:env Modeling}). The new GPU accelerated algorithm and implementation details are in Sec.~\ref{sec:GPU_Accelerate_Planning}. In Sec.~\ref{sec:Applications} we showcase the new algorithm by applying it to plan paths for different single-objective missions in a stochastic dynamic flow situation, clearly illustrating the computational efficiency of our algorithm. Finally, we conclude in Sec.~\ref{sec:conclusion} with directions for future research.



\subsection{Previous Progress in Optimal Path Planning} 
Most methods for optimal path planning of autonomous marine vehicles in stochastic dynamic environments are Monte Carlo versions of traditional deterministic planners (e.g., \cite{rathbun_et_al_2002,wang_hoteit_et_al_OD2016}) and surface response methods (e.g., \cite{kewlani_et_al_2009}). Unfortunately, they suffer from the curse of dimensionality and are computationally inefficient in serial implementation.
\cite{lermusiaux_et_al_TheSea2017,lermusiaux_et_al_springer_HOE2016} has a detailed review of deterministic planners such as dynamic Dijkstra \cite{mannarini_et_al_2016visir,garg2018dynamizing}, A$^*$ \cite{hart1968formal}, potential field functions \cite{barraquand_et_al_1992}, Delayed D$^*$ \cite{ferguson_stentz_2005ICRA}, Fast Marching Methods \cite{petres_et_al_2007}, nonlinear programming \cite{albarakati2020optimal}, and mixed integer linear programming \cite{yilmaz_et_al_JOE2008}.
\cite{smith2010autonomous, smith2010planning} use waypoint-selection algorithms and \cite{zeng2015efficient, zeng2020exploiting} use evolutionary algorithms like particle swarm optimization for path planning and utilize regular predictions from the regional ocean model system (ROMS). Recently, we have developed a suite of level-set based planners for deterministic environments \cite{lolla_et_al_OD2014_part1,subramani_lermusiaux_OM2016,kulkarni_lermusiaux_2020OM} and stochastic environments \cite{subramani_et_al_CMAME2018,subramani_lermusiaux_CMAME2019}. However, the stochastic versions give a distribution of paths and risk-optimal selection is a computationally expensive process.
The use of Markov Decision Processes (MDPs) for planning in stochastic environments \cite{alsabban2013wind, wolf2010probabilistic,kularatne_et_al_2018ICRA} has been proposed as they provide a rigorous mathematical framework for decision making under uncertainty. However, they become prohibitively expensive for large problems and use naive environment modelling techniques. For example, \cite{kularatne_et_al_2018ICRA} adds independent Gaussian noise to flow velocity components at each state to a mean forecast taken from ROMS, which causes the uncertainty to be spatially and temporally uncorrelated. Partially Observable Markov Decision Processes (POMDPs) have also been used for a variety of simultaneous localization and planning applications of ground or aerial vehicles where the focus is local closed-loop control \cite{du2011robot,amato_how_kaelbling_2017}.
Building on top of the MDP framework, reinforcement learning (RL) algorithms \cite{kober2013reinforcement,zhang2015geometric,finn2016guided,Yoo2016_RL,yijing2017q,singh2017methodology,chowdhury2020DDDAS} have also been used for path planning. 
GPU-accelerated RL libraries have also been developed recently \cite{babaeizadeh2016reinforcement},\cite{makoviychuk2021isaac}. 
However, RL algorithms provide approximate solutions to MDPs, which can be validated through exact dynamic programming solutions if the environment model is available and the solution is computationally feasible.
Other machine learning schemes such as Gaussian process regression \cite{sukhatme_GP_2018,mishra2018informed} have also been proposed.

The main drawback of the
MDP-based planners is the impractical computational time requirement of sequential implementation to solve large-scale problems. For example, \cite{kularatne_et_al_2018ICRA} reports that their MDP solution takes about 13 hours for a medium-scale autonomous underwater vehicles planning problem on a modern CPU. To accelerate the computation of optimal policies for large-scale problems, ideas of High-Performance Computing (HPC) have been proposed to solve the MDPs. For general problem statements, attempts have been made to parallelize the value iteration (VI) algorithm. \cite{johannsson2009gpu} provides a proof of concept where the value iteration algorithm was mapped to a GPU platform using CUDA. Value iteration was also implemented using OpenCL for path finding problems in \cite{chen2013markov}. \cite{herrera2014data} uses OpenMP and MPI to implement a very large-scale MDP to generate traffic control tables. \cite{ortega2019cuda} uses CUDA to implement value iteration on a GPU optimized for inventory control problems. \cite{ruiz2015parallel} fills state transition matrices and solves the MDP using a GPU in the context of crowd simulations, that lack a stochastic dynamic environment forecast. \cite{sapio2018efficient} implemented value iteration using CUDA, optimized for problems with sparse transition matrices. GPUs have also been used for other planning models such as semi-infinite nonlinear programming \cite{chretien_GPU_2016}. In the present paper, we develop an end-to-end GPU-based optimal path planning system that uses \textit{(i)} realistic forecasts of the ocean environment, \textit{(ii)} a newly developed GPU-implemented MDP model builder, and \textit{(iii)} an adaptation of the GPU-implemented value iteration algorithm \cite{sapio2018efficient} for solving the MDP. We also provide our software package for community use \cite{e2e_GPU_DP}.

\section{Markov Decision Process Framework}
\label{sec:MDPFramework}
Markov Decision Processes (MDPs) provide a natural framework for sequential decision-making to compute an optimal control law (policy) in stochastic environments. 
An MDP is defined as a tuple $<\mathcal{S}, \mathcal{A}, P, R,\gamma>$, where $\mathcal{S}$ is the set of states, $\mathcal{A}$ is the set of actions, $P$ denotes the state transition probabilities, $R$ denotes the one-step rewards, and $\gamma$ is the discount factor. A policy $\pi$ is defined as a map from the set of states $\mathcal{S}$ to a probability distribution over the set of actions $\mathcal{A}$. 
The state transition probabilities and and one-step rewards are $P(s'|s,a) = Pr(S_{t+1}=s'|S_t = s, A_t = a)$ and $R(s,a) = \mathbb{E}[R_{t+1}|S_t = s, A_t = a]$,
where, $S_{t}, A_{t}, R_{t}$ are random variables that denote the state, action and reward awarded at time $t$, and $s, s' \in \mathcal{S}, a \in \mathcal{A}$.
We define the value function $v_{\pi}(s)$ that quantifies the utility of following a policy $\pi$ from state $s$ as the expected sum of one-step rewards written in a recursive form in terms of the value function of possible successor states as 
\begin{linenomath}
\begin{align}
    v_{\pi}(s) &= \mathbb{E}_{\pi}\left[ \sum{R_t}|S_t=s\right] = \mathbb{E}_{\pi}\left[r + v_{\pi}(s')|S_t=s\right] \nonumber\\
                &=  \sum{P(s'|s,a)[r + v_{\pi}(s')] } \,.
 \end{align}
\end{linenomath}
An optimal policy $\pi_*$ is one that does better or equally as good as all the other policies $\pi'$, i.e., $v_{\pi_*}(s) \geq v_{\pi'}(s)$  $\forall~s~\in~\mathcal{S} $. The optimal value function $ v_*(s) = \max_{\pi}{v_{\pi}(s)}$ $\forall~s~\in~\mathcal{S}$ corresponding to an optimal policy can be expressed as a recursive relation (using Bellman conditions) 
\begin{linenomath}
\begin{align}
    v_*(s) =  \max_{a \in \mathcal{A}}\sum{P(s'|s,a)[r + v_*(s')] } \quad \forall s \in \mathcal{S}
    \label{eqn:bellman}
\end{align}
\end{linenomath}
that can be solved by dynamic programming algorithms. We use a finite horizon, undiscounted, total cost MDP for our path planning problem. Table.~\ref{tab:notation1} and Table.~\ref{tab:notation2} list all the notation used in the present paper.

\begin{table}
  \caption{Mathematical Notation and Symbols Used in the Present Paper}\label{tab:notation1}
  \setlength{\tabcolsep}{0.7\tabcolsep}
  \centering
  \resizebox{\columnwidth}{!}{%
  \begin{tabular}{ *{5}{l} }
    \toprule
    \textbf{Symbol} & \textbf{Description} \\
    \midrule
        $\mathcal{S}$ & Set of states for MDP. \\ 
        \hline
        $\mathcal{A}$ & Set of actions for MDP.\\
        \hline
        $P$ & State transition probabilities for MDP. \\
        \hline
        $R$ & One-step rewards for MDP.\\
        \hline
        $\mathcal{\gamma}$ & Discount factor for MDP.\\ 
        \hline
        $\pi$   & A policy for the MDP.\\
        \hline
        $\pi_{*}$ & An optimal policy for the MDP.\\
        \hline
        $v_{\pi}$ & Value function for the policy $\pi$.\\
        \hline
        $v_{*}$ & Optimal value function.\\
        \hline
        $v_{k}$ & $k^{th}$ update of the optimal value function in the value\\
        \quad & iteration algorithm.\\
        \hline
        $\mathbf{x}$ & 2D spatial coordinate representing agent's position. \\ 
        \hline
        $\mathbf{x}_s$  & Starting position of the agent. \\
        \hline
        $\mathbf{x}_f$ & Target position of the agent. \\
        \hline
        $\mathbf{x}_k$ & Position of the agent at timestep $k$. \\
        \hline
        $t$ &          Temporal coordinate. \\
        \hline
        $\omega_v$ & Sample of the stochastic velocity field. \\
        \hline
        $\omega_g$ & Sample of the stochastic scalar field. \\
        \hline
        $\mathbf{v}(\mathbf{x},t;\omega_v)$ & Stochastic dynamic velocity field. \\
        \hline
        $\bar{\mathbf{v}}(\mathbf{x},t)$ & Mean velocity field. \\
        \hline
        $\tilde{\mathbf{v}}_i(\mathbf{x},t)$& $i^{th}$ DO mode of the velocity field.\\
        \hline
        $\mu_i(t;\omega_v)$ & $i^{th}$ DO coefficient of the velocity field.\\
        \hline        
        $g(\mathbf{x},t;\omega_g)$ & Stochastic dynamic scalar field. \\
        \hline
        $\bar{g}(\mathbf{x},t)$ & Mean stochastic dynamic scalar field. \\
        \hline
        $m$ & Mask containing dynamic obstacle data.\\
        \hline
        $F$ & Speed of the agent with respect to currents. \\
        \hline
        $F_{max}$ & Maximum speed of the agent with respect to currents. \\
        \hline
        $f(.)$ & Agent's kinematic model. \\
        \hline
        $N_h$ & Number of possible headings for the agent. \\
        \hline
        $N_F$ & Number of possible speeds for the agent. \\
        \hline
        $N_{r,v}$ & Number of realizations of the stochastic velocity field. \\
        \hline
        $N_{r,g}$ & Number of realizations of the stochastic scalar field. \\
        \hline
        $N_m$ & Number of dominant modes of the stochastic velocity field.\\
        \hline
        $N_x$ & Dimension of the spatio-temporal grid along the $x$-axis. \\
        \hline     
        $N_y$ & Dimension of the spatio-temporal grid along the $y$-axis. \\
        \hline 
        $N_t$ & Dimension of the spatio-temporal grid along the $t$-axis. \\
        \hline 
        $N_c$ & Number of cells in the spatial grid at a given time. \\
        \hline
        $N_{sg}$ & Number of cells in a neighbouring sub-grid. \\
        \hline
        $N_g$ & Number of cells in the spatio-temporal grid. \\
        \hline
        $S_t$ & Random variable denoting the state of the agent at time $t$.\\
        \hline
        $A_t$ & Random variable denoting the action taken by the agent\\
        \quad & at time $t$.\\
        \hline
        $s$ & An element of the set $\mathcal{S}$. \\
        \hline
        $a$ & An element of the set $\mathcal{A}$.\\
        \hline
        $P(s'|s,a)$ & Probability that the agent transitions to a state $s'$ after\\
        \quad &  performing action $a$ in state $s$.  \\
        \hline
        $R(s,a,s')$ &  One-step reward given to the agent for reaching state $s'$ \\
        \quad &  upon performing action $a$ in state $s$ \\
        \hline
        $R(s,a)$ &  Expected one-step reward given to the agent upon \\
        \quad & performing action $a$ in state $s$. \\
        \hline        
        $r_{term}$ & A constant positive reward given to the agent for reaching\\
        \quad &  the terminal state. \\
        \hline
        $r_{outbound}$ & A constant negative reward given to the agent for going\\
        \quad &  outside the spatio-temporal domain or landing in an\\
        \quad & obstacle.\\
        \hline
        $e(.)$ & Energy function of the agent.\\
        \hline
        $c_r$ & Constant coefficient for scalar field terms.\\
        \hline
        $c_f$ & Constant coefficient in the energy function.\\
        \hline

    \bottomrule
  \end{tabular}%
  }
\end{table}

\subsection{MDP Formulation of Path Planning}
\label{problem_setup}
\begin{figure*}[h]
\centering
\includegraphics[width=\textwidth]{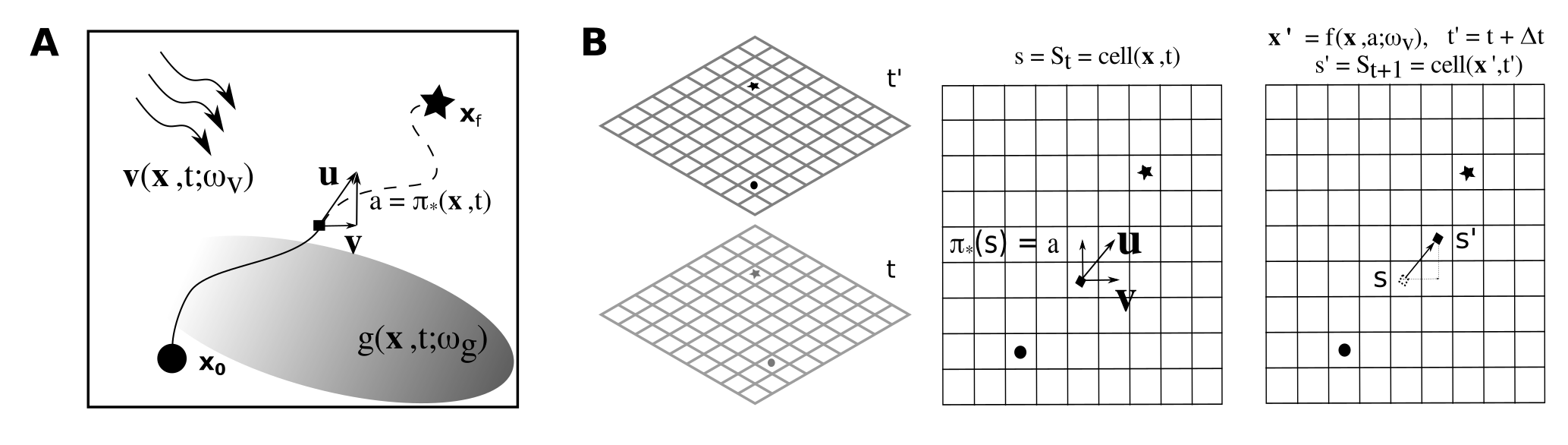}
\caption{\small \textbf{Schematic of the path planning problem:} \textbf{(A)} \textit{In the continuous state-space:} The autonomous agent must travel from $\mathbf{x}_0$ to $\mathbf{x}_f$ in a domain under the influence of a stochastic dynamic flow field $\mathbf{v}(\mathbf{x},t;\omega_v)$ and a stochastic dynamic scalar field $g(\mathbf{x},t;\omega_g)$ (e.g., solar radiation). The agent takes an action $a=\pi_*(\mathbf{x},t)$ according to an optimal deterministic policy $\pi_*$ and is simultaneously advected by an instantaneous velocity $\mathbf{v}$. The effective instantaneous path taken by the agent is along the vector sum $\mathbf{u=v}+a$. Meanwhile, it may also collect energy as it maneuvers through the scalar energy field $g$. We seek the optimal policy $\pi_*(\mathbf{x},t)$ that minimizes a given cost function based on the mission objective; \textbf{(B)} \textit{In the discrete state-space.} The discretized spatio-temporal grid corresponding to two successive time-steps $(t,t')$ are shown in the left panel. The middle and right panels show the spatial grid at $t$ and $t'$. At state $s$ (middle panel), the agent experiences a velocity $\mathbf{v}$, takes an action $a=\pi_*(s)$ and ends up in $s'$ in the next time-step (right panel). In doing so, the agent may also collect energy based on the radiations $g(s;\omega_g)$ and $g(s';\omega_g)$ at $s$ and $s'$, respectively. The agent receives a one-step reward,  $R(s,a,s';\omega_v, \omega_g$), upon making the transition.}
\label{Fig:schematic}
\end{figure*}

In a spatio-temporal domain $(\mathbf{x},t):\mathbb{R}^n\times[0,\infty)$, let $\mathbf{x}\in \mathbb{R}^n$ ($n=2,3$ for 2-D and 3-D space) denote the spatial index and $t \in [0,\infty)$ the temporal index (Fig.~\ref{Fig:schematic}A). Let the domain have a stochastic, dynamic flow $\mathbf{v}(\mathbf{x},t;\omega_v)$ where $\omega_v$ is a sample of the random velocity field with an associated probability distribution function. Let the domain also have a stochastic dynamic scalar field $g(\mathbf{x},t;\omega_g)$ (like solar, wind or wave energy field), where $\omega_g$ is a sample of the random scalar field. Let $F(t)$ be the instantaneous speed of an autonomous agent that starts from $\mathbf{x}_s$ at time $t=0$ and travels to $\mathbf{x}_f$ in a manner that minimizes the expected cost. 

Similar to the setup in \cite{chowdhury2020DDDAS} and \cite{kularatne_et_al_2018ICRA}, we discretize the domain to a spatio-temporal grid world 
that constitutes the state space (Fig \ref{Fig:schematic}(B)). A state $s$ represents a square region in the spatio-temporal grid, which is indexed by a spatio-temporal coordinate $[\mathbf{x_s},t_s]$. The square is centered at $\mathbf{x_s}$ with sides of length $\Delta{x}$. The grid is temporally discretized at intervals of $\Delta t$.
The random velocity at a discrete state is $\mathbf{v}(s;\omega_v)$. At each state $s$, the autonomous agent can take an action $a$ (heading and speed) from the action space $A$. We consider $N_h$ discrete headings between $0$ and $2\pi$, and $N_F$ discrete speeds between $0$ and $F_{max}$, where $F_{max}$ is the maximum possible relative speed of the agent. Hence the action space consists of $|\mathcal{A}| = N_h*N_F$ discrete actions. Our problem is to find the optimum deterministic policy $\pi_*(s)$ that the agent must follow to minimize the expected total cost of the mission.

The state transition probabilities $P(s'|s,a)$ denotes the probability that the agent lands in the state $s'$, given that it performs an action $a$ in state $s$. This probability is induced from the uncertainty in the velocity field, $\mathbf{v}(s;\omega_v)$ through the vehicle kinematics. The spatial motion is given by $\mathbf{x'} = f(\mathbf{x}, a; \omega_v) = \mathbf{x} + (\mathbf{v}(s;\omega_v) + a)\Delta t$, and the temporal update is $t'=t+\Delta t$. The successor state is computed as $s' = cell(\mathbf{x'},t')$, where the $cell()$ function maps the spatial and temporal coordinates to the appropriate discrete state in the domain. When the uncertainty in the environment is not in a simple closed-form (e.g., Gaussian), we often have to use a Monte-Carlo based approach \cite{wolf2010probabilistic}, i.e., enumerating multiple samples of the uncertain field and counting how many times a successor state $s'$ is visited under the influence of action $a$ performed in a given state $s$. 

The expected one-step rewards $R(s,a)$ that an agent receives upon reaching a state $s'$ after taking an action $a$ in state $s$ depends on the mission objective. For the three different mission objectives, we consider they are defined as follows. (i) \textit{Minimize expected travel time}: The agent is penalized a constant value for each time-step leading to the one-step reward 
\begin{linenomath}
    \begin{equation}
        R(s,a) = -\Delta{t}\,,
    \label{eqn:Rsa_time}
    \end{equation}
\end{linenomath}
	(ii)
	\textit{Minimize expected energy consumption}: The agent's energy consumption is modeled as a function of its speed leading to the one-step reward 
	\begin{linenomath}
    \begin{equation}
        R(s,a) = -e(F)\Delta t\,,
    \label{eqn:Rsa_energyConsumption}
    \end{equation}
    \end{linenomath}
    where $F$ is the agent's speed and $e$ is the energy function. Typically, a quadratic relationship of the form $e(F)~=~{c_f F^2}$ is utilized, where $c_f$ is a constant \cite{subramani_lermusiaux_OM2016}; and 
    (iii) \textit{Minimize expected net energy consumption}: The agent is equipped with a solar panel (or any other energy harvesting mechanism) that collects energy from a spatio-temporal stochastic dynamic energy field $g(\bullet)$ while moving around in the domain. Hence, the expected one-step reward is defined as
    \begin{linenomath}
    \begin{equation}
        R(s,a) = \mathop{\mathbb{E}}\left[-e(F) + c_r\left(\frac{g(s) + g(s')}{2}\right)\right]\Delta t\,,
    \label{eqn:Rsa_NetEnergy}
    \end{equation}
    \end{linenomath}
    where $c_r$ is a constant. The first term inside the expectation $-e(F)$ is the energy consumed for taking the action $a$. We assume that the energy harvested by the agent is directly proportional to the magnitude of the solar radiation field. The second term is a first-order integral approximation of the energy collected by the agent from $g(\bullet)$ as it transitions from the state $s$ to $s’$. The sum of the two terms is the agent's one-step net-energy consumption. The expectation in Eq.~(\ref{eqn:Rsa_NetEnergy}) is taken over the joint distribution $P(s',\omega_g|s,a)$. Moreover, irrespective of the mission objective, the agent is given a positive reward of $r_{term}$ if it reaches the target terminal cell to provide it an incentive to reach the target location. In the absence of this incentive, the agent may never reach the target location. Similarly, it is given a relatively large negative reward $r_{outbound}$ if it lands in a state marked as an obstacle, goes out of the defined spatial domain, or requires time beyond the finite planning horizon. Note that here we emphasize a 2D in space problem setup; however, our framework and implementation apply to 3D in space as well. Also, we use a kinematic framework for planning, without considering the dynamics of the flow around the UAV \cite{lolla_et_al_OD2014_part1,alsabban2013wind,wolf2010probabilistic,kularatne_et_al_2018ICRA,chowdhury2020DDDAS}.

\subsection{Dynamic Programming Algorithm}
\label{subsec:DP}
Since our planning problem is a finite horizon, total cost, undiscounted MDP, there exists an optimal deterministic policy $\pi: \mathcal{S} \xrightarrow[]{} \mathcal{A}$ \cite{bellman1957dynamic} that can be computed using dynamic programming. The key idea of dynamic programming is to use the value functions to organise and structure the search for good policies. Value iteration can be viewed as turning the Bellman optimality Eq.~(\ref{eqn:bellman}) into the update rule
\begin{linenomath}
 \begin{align}
    v_{k+1}(s) =  \max_{a \in \mathcal{A}}\sum{P(s'|s,a)[r + v_{k}(s')] }\quad \forall\, s \in \mathcal{S}\,.
\end{align}
 The sequence ${v_k}$ can be shown to converge to the optimal $v_*$ for any arbitrary initial value $v_0$, under the same conditions that guarantee the existence of $v_*$ \cite{bertsekas1995dynamic}. Once the optimal value function is obtained, the optimal policy can directly be computed as  
 \begin{align}
     \pi_*(s) = \argmax_{a \in \mathcal{A}}{\sum{P(s'|s,a)[r + v_{*}(s')] }} \quad \forall\, s \in \mathcal{S}\,.
 \end{align}
\end{linenomath}
We refer the reader to \cite{bertsekas1995dynamic} for the complete algorithm.

\subsection{Stochastic Dynamic Environment Modeling}
\label{sec:env Modeling}
A crucial input for computing the state transition probabilities of the MDP path planning problem is the stochastic dynamic environmental flow field $\mathbf{v}(\mathbf{x},t;\omega_v)$ in its decomposed form $\mathbf{v}(\mathbf{x},t;\omega_v)=\bar{\mathbf{v}}(\mathbf{x},t)+\mu_i(t;\omega_v)\tilde{\mathbf{v}}_i(\mathbf{x},t)$, where $\bar{\mathbf{v}}(\mathbf{x},t)$ is the so-called DO mean, $\tilde{\mathbf{v}}_i(\mathbf{x},t)$ are the DO modes, and $\mu_i(t;\omega_v)$ are the DO coefficients. Here, Einstein's notation (repeated indices indicate summation) is used. In the present paper, we employ the computationally efficient Dynamically Orthogonal field equations \cite{sapsis_lermusiaux_PD2009}, in which the DO mean, modes and coefficients are solved by integrating explicit equations for them \cite{subramani_et_al_CMAME2018,chowdhury2020DDDAS}. Alternatively, any modeling system that provides the flow field's distribution in terms of the above decomposition is sufficient to compute the transition probability matrix and complete the planning. For example, a deterministic ocean modeling system (e.g., \cite{haley_lermusiaux_OD2010,shchepetkin2005regional,nemo_book}) may be employed to forecast an ensemble from which the dynamically orthogonal reduced-order representation can be obtained by performing a singular value decomposition \cite{lermusiaux_robinson_MWR1999}. Further, any other uncertainty quantification and propagation method such as the Polynomial Chaos Expansion \cite{ghanem_spanos_2003} can also be used. Notably, the DO equations provide a significant computational advantage (two to four orders of magnitude speedup) over the ensemble approach. Similarly, the stochastic dynamic scalar field $g(\mathbf{x},t;\omega_g)$ representing the energy field to be collected by the vehicle must be obtained using environmental models such as atmospheric \cite{wrf_manual_2019}, wind or wave \cite{wavewatch_manual_2019} models. 


\section{GPU Accelerated Planning}
\label{sec:GPU_Accelerate_Planning}
There are two phases in the algorithm to compute an optimal policy of the MDP:
\textit{(1)}	Building the model, i.e., computing the expected one-step rewards $R$, and the state transition probabilities $P$;
\textit{(2)}	Solving the MDP, i.e., obtaining an optimal policy using methods such as dynamic or linear programming.
Each of these phases requires practically infeasible computational time in serial implementation. The cost of the first phase using a naive Monte-Carlo approach is $O(N_{r,v}N_{r,g}|\mathcal{A}||\mathcal{S}|)$ as we must compute the one-step transition data for each state, for each action, and for each realization of the uncertain velocity and scalar fields. Here, $N_{r,g}$ is the number of realizations of the stochastic scalar field $g(s;\omega_g)$. The cost of the second phase using value iteration is $O(n|\mathcal{A}||\mathcal{S}|^2)$. Therefore, when we have a large state-, action-, or realization-space, the compute time may become impractical. To solve large-scale problems, without having to resort to approximate methods such as Reinforcement Learning \cite{sutton2018reinforcement}, we propose a GPU-based algorithm to achieve significant computational speedups. Since our algorithm provides a dynamic programming solution, it can be used to validate solutions obtained through reinforcement learning methods. Next, we discuss each phase separately, wherein we identify the useful properties that make them suitable for parallelization and how they can be implemented on a GPU. We also discuss how to overcome the physical device limitations of a GPU through algorithmic intervention.


\subsection{MDP Model Building on GPUs}
\label{subsec:MDP Model building}
To compute the transition probabilities $P$ and the expected one-step rewards $R$ for the MDP, we must calculate one-step transitions for each state, each action, and each realization by evaluating $R(s_i,a_j;\omega_{v_k})$ and $f(s_i,a_j;\omega_{v_k})$. A key observation in the process is that these computations are independent of each other across states, actions and realizations, i.e., $R$ and $f$ for any $(i,j,k)$ tuple can be computed independent of any other $(i,j,k)$ tuple.

Hence, the key idea is that one can spawn thousands of independent threads that perform all of these computations for every $(i,j,k)$ tuple independently and in parallel. Additionally, since there is no need for any data communication between these parallelly spawned threads, the application is ``embarrassingly parallel". Such applications are ideal for being implemented on a GPU. In our work, we use NVIDIA GPUs, and the algorithm is written using CUDA.

Ideally, with a sufficient number of cores, all the computation can potentially happen simultaneously in parallel. However, practically, we would like our algorithm to run efficiently on modest GPUs installed on the autonomous system itself and not consume excessive energy (to operate the GPU). Our ultimate goal is to update the MDP model through in-mission measurements of the surrounding environment and update the optimal policy regularly with a modest on-board GPU. Towards this goal, in the present paper, we focus only on the end-to-end GPU algorithm for solving the offline path-planning problem. Critically, such modest GPUs may have resource limitations such as limited memory. For instance, if we decided to store the velocity field of the entire spatio-temporal grid of size $N_{g}=N_x\times N_y\times N_t$ (say $100\times 100\times 100$) with $N_{r,v}$ (say 1000) realizations as floating points  in GPU memory, it would require 8 GB of global memory! Clearly, this is not acceptable. Hence, we propose the following three solutions to deal with such limitations.

First, we use a reduced-order velocity field that is stored in the GPU's global memory. The $N_{r,v}$ realizations of the stochastic, dynamic velocity field are decomposed to a mean-field along with corresponding modes and coefficients. For instance, let’s assume that the first $N_m$ (say 10) singular values of the velocity's covariance matrix capture a specified threshold of the variability (say 99\%), then for the aforementioned spatio-temporal grid and realizations, each component of the velocity field would be broken down to a mean-field matrix of size $N_{g}$ $(100\times 100\times 100)$, $N_m$ mode matrices of size $N_g$ $(100\times 100\times 100)$ each, and a coefficient matrix of size $N_{m}\times N_{r,v} \times N_t$ ($10\times 1000\times 100$). This requires only 96 MB of memory compared to the 8 GB required otherwise, which is roughly a two orders of magnitude reduction.

Second, we use just the mean scalar field for computing the expected one-step rewards when the problem involves the agent manoeuvring to collect energy from the stochastic scalar field. To do so, we show that $N_{r,g}$ enumerated samples of the field are not required to compute the expected one-step rewards, and just the mean-field is sufficient. The use of only the mean scalar field helps save memory and saves computational expense by a factor of $N_{r,g}$. The proof is as follows. The optimal value function can be expressed in terms of an expectation for all $s \in \mathcal{S}$ as
\begin{linenomath}
    \begin{align}
    v_*(s) &= \max_{a \in \mathcal{A}}\mathbb{E}[R(s,a,s') + v_*(s')] \label{v*s=E[]}\,,\\
    R(s,a,s') &= \left[-c_fF^2 + c_r\left(\frac{g(s;\omega_g) + g(s';\omega_g)}{2}\right)\right]\Delta t \,,\\
    R(s,a) &= \mathbb{E} R(s,a,s')\,,\\
    &= \mathbb{E}\left[-c_fF^2 + c_r\left(\frac{g(s;\omega_g) + g(s';\omega_g)}{2}\right) \right]\Delta t\label{eqn:R(s,a)=E1}\,,\\
    \begin{split}
    &= - c_fF^2 \Delta t + c_r\mathbb{E}\left[\frac{g(s;\omega_g)}{2}\right]\Delta t  \\ 
    &\qquad + c_r\mathbb{E}\left[\frac{g(s';\omega_g)}{2}\right]\Delta t \label{eqn:R(s,a)=E2}\,.
    \end{split}
    \end{align}
\end{linenomath}
Let us focus on the second and third energy terms on the RHS of Eq.~(\ref{eqn:R(s,a)=E2}). In the second term, the only random variable inside the expectation is $\omega_g$. Note that $s$ is not a random variable in this equation since we are calculating the expected one-step reward for performing an action $a$ in a given state $s$, which is known. Therefore, the expectation in the second term is over the probability distribution $P(\omega_g|s,a)$ or simply $P(\omega_g|s)$, since $\omega_g$ is independent of the action $a$. The expectation simply results in the mean value of $g$ at state $s$ (Eq.~\ref{eqn:Eg(s,ls)=gbars}). However, in the third term, both $s'$ and $\omega_g$ are random variables. Hence this expectation is over the joint distribution $P(s',\omega_g|s,a)$ and can be expressed as an iterated conditional expectation (Eq.~\ref{eqn:EEg(s,l)}). The outer expectation is over the distribution $P(s'|s,a)$, which are the state transition probabilities. The inner expectation is over the distribution $P(\omega_g|s',s,a)$, which simplifies to $P(\omega_g|s')$ since $\omega_g$ is independent of $s$ and $a$ when evaluating $g$ at $s'$ (Eq.~\ref{eqn:EEEg(s,l)_2}). The inner expectation then simply results in mean value of $g$ at $s'$ (Eq.~(\ref{eqn:E[gs',ls']=Egbars'})).
\begin{linenomath}
    \begin{align}
        \mathbb{E}\left[g(s;\omega_g)\right] &= {\mathbb{E}}_{\omega_g|s}\left[g(s;\omega_g)\right] = \overline{g}(s) \label{eqn:Eg(s,ls)=gbars}\,,\\
        \mathbb{E}\left[g(s';\omega_g)\right]&={\mathbb{E}}_{s'|s,a}\left[{\mathbb{E}}_{\omega_g|s',s,a}\left[g(s';\omega_g)\right]\right] \label{eqn:EEg(s,l)} \\
        &={\mathbb{E}}_{s'|s,a}\left[{\mathbb{E}}_{\omega_g|s'}\left[g(s';\omega_g)\right]\right]\label{eqn:EEEg(s,l)_2}\\
        &=  {\mathbb{E}}_{s'|s,a}\left[\overline{g}(s')\right]\,.\label{eqn:E[gs',ls']=Egbars'}
    \end{align}
\end{linenomath}
From Eq.~(\ref{eqn:Eg(s,ls)=gbars}) and Eq.~(\ref{eqn:E[gs',ls']=Egbars'}) we can see that the second and third terms in the RHS of Eq.~(\ref{eqn:R(s,a)=E2}) depend only on $\overline{g}(s)$ and $\overline{g}(s')$, which are the means of the stochastic scalar field at $s$ and $s'$, respectively. This holds true for any $s$ and $s'$. Hence, while we may have a model characterising the uncertainty in the scalar field and it is possible to take $N_{r,g}$ samples of the same, it suffices to store only the mean scalar field, $\overline{g}$. Finally, substituting Eq.~\ref{eqn:Eg(s,ls)=gbars} and Eq.~\ref{eqn:E[gs',ls']=Egbars'} in Eq.~\ref{eqn:R(s,a)=E2}, we get 
\begin{linenomath}
    \begin{align}
           R(s,a) &= \left[-c_fF^2 + \frac{1}{2}c_r\overline{g}(s) + \frac{1}{2}c_r{\mathbb{E}}_{s'|s,a}\left[\overline{g}(s')\right]\right]\Delta t \\
             &={\mathbb{E}}_{s'|s,a}\left[ - c_fF^2 + \frac{1}{2}c_r\overline{g}(s) + \frac{1}{2}c_r\overline{g}(s') \right]\Delta t
    \end{align}
\end{linenomath}
Third, we use the concept of a secondary grid, which we refer to as the neighbouring sub-grid, whose size is much smaller than the spatial grid at a given time instance. For a given $(s, a)$ pair, the neighbouring sub-grid is used to store the possible values of $s'$ for different realizations in a memory-efficient manner. The algorithm and the details of how the neighbouring sub-grid achieves memory efficiency are discussed next.

\subsection{CUDA Algorithm for Building the MDP Model}\label{subsec:algo}

\begin{singlespace}
\begin{table}[h]
  \caption{Mathematical Notation and Symbols Used in Algorithms in the Present Paper}\label{tab:notation2}
  \centering
  \resizebox{\columnwidth}{!}{%
  \begin{tabular}{ *{5}{l} }
    \toprule
    \textbf{Symbol} & \textbf{Description} \\
    \midrule
        $S2\_count$ & An array that keeps a count of the number of times a \\
        \quad & successor state $s'$ is reached upon taking an action $a$ \\
        \quad & for each initial state $s$, across $N_{r,v}$ realizations.\\ 
        \hline
        $sum\_R$ &  An array that stores the sum of one-step rewards \\
        \quad & across $N_{r,v}$ realizations for each initial state $s$,\\
        \quad & for a given action $a$ at a given time step.\\
        \hline
        $P_{a,t}$ & A COO format state transition matrix for transitions\\
        \quad &  from state $s$ at time $t$ to the states $s'$ at \\
        \quad & the successive time $t'$, for the given action $a$.\\
        \hline
        $R_{a,t}$ & Reward vector that stores the expected one-step rewards \\
        \quad & for performing the action $a$ in each state at time $t$.\\
        \hline
        $nnz$ & Number of non-zero elements in $P_{a,t}$.\\ 
        \hline
        $nnz\_count$ & An array that stores the number of possible successor \\
        \quad & states $s'$ for for each state $s$ at time $t$.\\ 
        \hline
        $P_a$  & A COO format state transition matrix for all states \\
        \quad & for the given action $a$.\\
        \hline
        $R_a$ & Reward vector that stores the expected one-step rewards \\
        \quad & for all states for performing the given action $a$.\\
        \hline
        $\underbar{P}$ & A COO format state transition matrix made by appending \\
        \quad & $P_a$s for all actions $a$.\\
        \hline
        $\underbar{R}$ & Reward vector containing expected one-step rewards\\
        \quad & made by appending $R_a$s for all actions $a$.\\
        \hline
        $tid$ & Thread index assigned to each thread in a kernel. \\
        \hline
        $s\_id_{x,y}$ & A one-dimensional index to denote discrete cell \\
        \quad & in a spatial grid at any given time.\\
        \hline
        $s\_id_{x,y,t}$ & A one-dimensional index to denote discrete cell \\
        \quad & in the spatio-temporal grid world.\\
        \hline
        $\omega\_id$ & A one-dimensional index to denote the realization of \\
        \quad & the stochastic velocity field $\mathbf{v}$.\\ 
        \hline
        $g_s$ & Mean scalar field at state $s$.\\
        \hline
        $r$ & One-step reward.\\
        \hline
        $id1$ & An index to access $S2\_count$ array.\\
        \hline
        $id2$ & An index to access $sum\_R$ array.\\
        \hline
    \bottomrule
  \end{tabular}%
  }
\end{table}
\end{singlespace}

Algorithm \ref{algo:building the model} is the newly developed CUDA algorithm for building the MDP transition probability matrix and the expected one-step reward.

\begin{algorithm}

\begin{algorithmic}[1]
\SetAlgoLined
    \REQUIRE{$\bar{\mathbf{v}},\tilde{\mathbf{v}}_i,\mu_i ,\overline{g},m,prob\_type, prob\_params,grid\_params$}
    \ENSURE{$\underline{P}, \underline{R}$}
    \STATE MEMCPY $\bar{\mathbf{v}},\tilde{\mathbf{v}}_i, \mu_i , \overline{g} $ to GPU\;
    \STATE Compute dimension for neighbouring sub-grid\;
    \STATE Allocate memory for $S2\_count$ of size $N_{sg}N_c$\;
    \STATE Allocate memory for $sum\_R$ of size $N_c$\;
    \FOR{($t=0;t<N_{t};t++$)}
        \FOR{ $a\in A$ }
        \STATE $S2\_count$, $sum\_R \gets  K\_TRANSITION(t, a, \overline{\mathbf{v}}, \tilde{\mathbf{v}}_i, \mu_i, \overline{g}, m)$\;
        \STATE $R_{a,t} \gets K\_COMPUTE\_MEAN(sum\_R)$\;
        \STATE $nnz, nnz\_count \gets  K\_COUNT(S2\_count)$\;
        \STATE $P_{a,t} \gets Allocate\_memory\_for\_COO(nnz)$\;
        \STATE $P_{a,t} \gets K\_REDUCE\_TO\_COO(P_{a,t},nnz\_count)$\;
        \STATE $MEMCPY\_DEV\_TO\_HOST(P_{a,t})$\;
        \STATE $P_{a}$, $R_{a} \gets  Append\_model\_in\_time(P_{a,t}, R_{a,t})$\;
        \ENDFOR
    \ENDFOR
 \FOR{ $a\in A$ }
    \STATE $\underline{P}, \underline{R} \gets  Append\_model\_in\_actions(P_{a}, R_{a})$\;
 \ENDFOR
 
 \end{algorithmic}
 \caption{Building the MDP Model}
 \label{algo:building the model}
\end{algorithm}

Algorithm \ref{algo:building the model} takes the following input,
\begin{enumerate}
    \item 
    \textit{Environment data}: Environment data comprises all environment information such as the reduced-order stochastic velocity field in the form of a mean velocity field, $\bar{\mathbf{v}}$, the dominant modes, $\tilde{\mathbf{v}}_i$, and corresponding coefficients, $\mu_i $; the mean of the stochastic scalar field, $\overline{g}$; and the dynamic obstacle information in the form of a mask, $m$.
    
    \item
    $prob\_type$: A parameter that defines the problem type and objective function. The objective could be minimizing expected travel time, expected energy consumption or expected net energy consumption.
    
    \item
    $prob\_params$: An array that stores other problem parameters such as $F_{max}$, $N_h$, $N_F$, which are used to define the action space, and $r_{outbound}$ and $r_{term}$, which are used to define the reward structure.
    
    \item 
    $grid\_params$: An array that stores grid parameters such as length scales, time scales, and sizes of the spatio-temporal grid, $N_x$, $N_y$ and $N_t$.
    
\end{enumerate}
and gives $\underline{P}$ and $\underline{R}$ as the output, where  $\underline{P}$ is an appended transition matrix and $\underline{R}$ is the appended expected rewards vector.

\begin{figure*}[h]
\centering
\includegraphics[width=\textwidth]{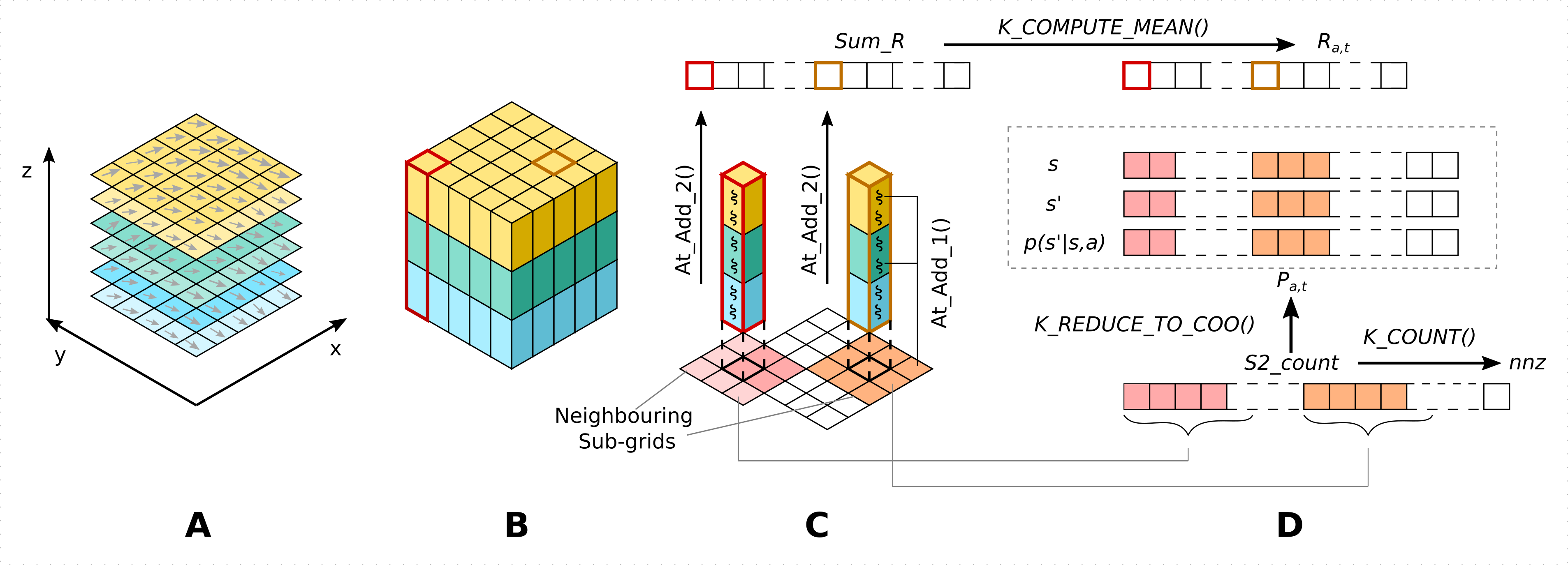}
\caption{\small \textit{Methodology for building the MDP model using a GPU}: \textbf{(A)} A 3D grid showing various realizations of the stochastic velocity field at a given time-step. \textbf{(B)} A 3D grid of CUDA threads, mimicking the structure of the 3D grid in (A). \textbf{(C)} Example showing participation of thread blocks during the execution of the  $K\_TRANSITION$ kernel for two different spatial coordinates at a given time-step for a given action, i.e., for two different $(s,a)$ pairs. The figure also shows two neighbouring sub-grids highlighted in red and orange.
\textbf{(D)} The information in neighbouring sub-grids is written to $S2\_count$ array. The $K\_COUNT$ kernel computes the number of non-zero elements, followed by the $K\_REDUCE\_TO\_COO$ kernel, which converts the information to a sparse COO-format transition probability matrix. On the other hand, the $K\_COMPUTE\_MEAN$ kernel computes the expected reward vector $R_{a,t}$ from the $sum\_R$ array.  }
\label{fig:methodology}
\end{figure*}

The algorithm works as follows. First, the input data, which consists of environmental information, the grid, and problem parameters are transferred from the CPU to the GPU using a \textit{MEMCPY} operation so that it can be readily accessed by threads during kernel launches. Next, we define the size of the neighbouring sub-grid. Conceptually, the neighbouring sub-grid is a smaller, rectangular 2D grid defined for each state at a given time-step in the domain. The key idea behind the grid is that an agent starting at a state $s$ cannot spatially transition to any other state $s'$ that lies outside the neighbouring sub-grid of state $s$, since it can only transition to physically nearby states. Fig.~\ref{fig:methodology}(C) shows neighbouring sub-grids for two different initial states. The size of the sub-grid ($N_{sg})$ is kept constant across all states and depends on the component-wise maximum velocities across all states with an added buffer.
The cells inside the neighbouring sub-grid map to elements in $S2\_count$ array as shown in Fig.~\ref{fig:methodology}(C) and Fig.~\ref{fig:methodology}(D). The details of this mapping and how the sub-grid aids efficient computation are discussed in \ref{subsubsec:K_one_step}.

After that, we allocate memory for the $S2\_count$ and $sum\_R$ arrays in the GPU. The size of $S2\_count$ array is $N_{sg}N_c$, where $N_c$ is the number of cells (or states) at a given time step. The $S2\_count$ array keeps a count of the number of times a given successor state $s'$ is reached upon taking a given action $a$ for each initial state $s$, across $N_{r,v}$ realizations. On the other hand, the size of the $sum\_R$ array is $N_c$. It stores the sum of one-step rewards across $N_{r,v}$ realizations for each initial state $s$, for a given action $a$ at a given time step. These arrays are filled in by successive kernel calls described in Sec. \ref{subsec:Kernels}.

Next in the algorithm, we have a series of CUDA kernels responsible for computing the MDP model. These kernels are nested within two for-loops - the outer loop ranges across time-steps, and the inner loop ranges across all actions in the action space. Hypothetically, if the GPU had unlimited memory and compute cores, all the computation could potentially be done in parallel without iterating over time-steps and actions. However, modest GPUs have limited memory and compute cores, making it difficult, if not impossible, to not only store the data but also be able to perform all computations parallelly. 
Since all computations cannot happen in parallel, splitting the computation through loops does not compromise maximum achievable parallelism. Effectively, we will be using the GPU to its full potential while splitting computations across time-steps and actions. As such, each kernel performs computations over states at a given time-step and a given action. In what follows, we discuss each kernel's working and implementation.

\subsection{CUDA Kernels}
\label{subsec:Kernels}

\subsubsection{K\_TRANSITION}
\label{subsubsec:K_one_step}

\begin{algorithm}
\begin{algorithmic}[1]
    \SetAlgoLined
    \REQUIRE $t, a, \overline{\mathbf{v}}, \tilde{\mathbf{v}}_i, \mu_i, \overline{g}, m$
    \ENSURE $ S2\_count, sum\_R$
     
        \STATE $tid \gets get\_thread\_id()$\;
        \STATE $s\_id_{x,y}, s\_id_{x,y,t}, \omega\_id \gets get\_ids(t, tid) $\; \label{algline:ids}
        \IF{($tid<N_{x}N_{y}N_{r,v}$)}
            \STATE $v_x, v_y \gets compute\_velocity(s\_id_{x,y,t}, \omega\_id, \mathbf{v}, \tilde{\mathbf{v}}_i, \mu_i)$\; \label{algline:vx}
             \STATE $g_s \gets read\_scalar\_field( s\_id_{x,y,t},  \overline{g}$)\;\label{algline:g_s}
            \IF{($s\_id_{x,y,t}$ is not terminal)}\label{algline:if_not_term}
                \IF{($s\_id_{x,y,t}$ is not obstacle)}\label{algline:if_not_obs}
                     \STATE $s\_id_{x',y'}, s\_id_{x',y',t'} \gets move(a, v_x, v_y$)\;\label{algline:move}
                     \STATE $g_{s'} \gets read\_scalar\_field(s\_id_{x',y',t'},  \overline{g}$)\;\label{algline:g_s'}
                     \STATE$r \gets r + compute\_reward (prob\_type, r_{term}, r_{outbound} $)\;\label{algline:onestep_r}
                    \IF{(obstacle between $s\_id_{x',y'}$ and  $s\_id_{x',y',t'}$)}\label{algline:if_goes_thru}
                         \STATE $r \gets r_{outbound}$\;
                    \ENDIF
                \ENDIF
            \ELSE   \label{algline:else}
                \STATE  $r \gets r_{outbound}$\;
            \ENDIF
            \STATE $id1 \gets get\_S2\_array\_idx()$\;\label{algline:id1}
            \STATE $id2 \gets get\_sumR\_idx($)\;\label{algline:id2}
            \STATE $S2\_count \gets atomicAdd(\&S2\_count[id1], 1$)\;\label{algline:atAdd1}
            \STATE $sum\_R \gets atomicAdd(\&sum\_R[id2], r$)\;\label{algline:atAdd2}
       \ENDIF
\end{algorithmic}
 \caption{K\_TRANSITION}
 \label{algo:K_one_step_transition}
\end{algorithm}

This is the most important kernel in the algorithm and consumes the majority of the total compute time. The kernel's purpose is to compute one-step transitions for each state at a given time-step and action across all $N_{r,v}$ realizations of the velocity field. The computed data is then written to the appropriate indices of the $S2\_count$ and $sum\_R$ arrays. Algorithm~\ref{algo:K_one_step_transition} shows the pseudo-code of the kernel, which is executed parallelly by all the threads. 
A visual representation of the stochastic velocity field at a given time-step is shown in Fig.~\ref{fig:methodology}(A) in the form of a 3D grid. Each plane in the grid represents the spatial domain marked with grey arrows, highlighting the velocity field corresponding to a realization at a given time-step. The field's $N_{r,v}$ realizations are stacked along the $z$-direction.
The kernel is launched using a 3D grid of thread blocks (Fig.~\ref{fig:methodology}(B)) that mimics the structure of the 3D grid of velocity field realizations in Fig.~\ref{fig:methodology}(A). The $x$ and $y$ directions of the kernel's grid correspond to the $x$ and $y$ directions of the spatial domain, and the $z$-direction corresponds to the $N_{r,v}$ realizations of the velocity field. All thread blocks (cuboidal units of the kernel's grid) that share the same $x$-$y$ coordinate are stacked along the $z$-direction and map to the corresponding $x$-$y$ coordinate on the spatial domain. For example, one such stack outlined in red corresponds to the spatial cell in the left corner. Also, each thread block consists of a single queue of threads oriented along the $z$-direction such that each thread in the block maps to a realization of the velocity field.  Fig. \ref{fig:methodology}(C) shows two stacks of blocks that correspond to a given state, with each block consisting of multiple threads along the $z$-direction, corresponding to the velocity field realizations (Fig.~\ref{fig:methodology}(A)).  
As a result of this mimicking structure,
each thread uniquely maps to a particular state and realization at the given time-step, 
and maintains one-dimensional indices corresponding to \textit{(i)} the discrete cell in the spatio-temporal grid world denoted by $s\_id_{x,y,t}$,  \textit{(ii)} the discrete cell in the spatial grid at time $t$, denoted by $s\_id_{x,y}$, and  \textit{(iii)} realization of the velocity field, denoted by $\omega\_id$ (Line~\ref{algline:ids} of Algorithm~\ref{algo:K_one_step_transition}). Therefore, each thread can independently compute the velocity field's and scalar field's numerical values for its corresponding state and realization from the mean, modes, coefficients and mean scalar field, already present in the GPU's global memory (Lines \ref{algline:vx} and \ref{algline:g_s}). Similarly, each thread can also independently access the obstacles data from the mask matrix $m$.
Once a thread has computed the environment information for its state and realization, it independently computes $s' = cell(\mathbf{x'},t')$, i.e., the successor state's spatial coordinate $\mathbf{x'} = f(\mathbf{x},a,\omega_v)$ and temporal coordinate $t' = t + \Delta t$, through the $move()$ function in Line \ref{algline:move}.
Next, since each thread knows its $s, a, s', \omega_v$ values, it reads the value of the mean field at $s'$ and then computes its corresponding one-step reward $R(s, a, s'; \omega_v)$ using the $compute\_reward()$ function (Lines \ref{algline:g_s'}, \ref{algline:onestep_r}), wherein  the reward structure is defined according to Eq.~\ref{eqn:Rsa_time}-\ref{eqn:Rsa_NetEnergy}. Computing the one-step reward also involves checking for the presence of any obstacles between $s$ and $s'$ and whether $s'$ is a terminal state by evaluating the if-conditions in lines \ref{algline:if_not_term}, \ref{algline:if_not_obs}, and \ref{algline:if_goes_thru}.

Based on our definition of the neighbouring sub-grid (Sec.~\ref{subsec:algo}), the successor state $s'$ must be a part of the sub-grid itself. Fig.~\ref{fig:methodology}(C) shows an example of neighbouring sub-grids for two different initial states, highlighted in red and orange colours. 
Each cell in the neighbouring sub-grid 
maps to a contiguous set of $N_{sg}$ elements in the $S2\_count$ array, as shown in Fig.~\ref{fig:methodology}(C) and Fig.~\ref{fig:methodology}(D). The $S2\_count$ array is of size $N_{sg}\times{N_c}$ since it maps from $N_{sg}$ sized neighbouring sub-grids for each of the $N_c$ possible initial states. 
Each thread computes the sub-grid index corresponding to $s'$ and the corresponding index $id1$ (Line~\ref{algline:id1}) to access the $S2\_count$ array. Thereafter, it performs an $atomicAdd()$ operation (Line~\ref{algline:atAdd1}), incrementing the count at the index $id1$ of the $S2\_count$ array. Fig.~\ref{fig:methodology}(C) provides a visual representation of the above process through a representative stack of three thread blocks (six threads in total) highlighted in orange. It shows how threads with the same sub-grid index participate in an $atomicAdd()$ operation named $At\_add\_1()$ to access and write to the appropriate element in the $S2\_count$ array. In this way, the $S2\_count$ array encodes the number of times an agent starting at an initial state $s$ and performing a given action $a$ transitions to a successor state $s'$ for all possible initial and successor states at times $t$ and $t'$, respectively. 

Crucially, we note that the neighbouring sub-grid concept limits the size of the $S2\_count$ array by using the fact that transitions can practically only happen to nearby cells. The naive alternative would be to allow parallel access to $S2\_count$ array by making it of the size $N_c\times{N_c}$. The advantage of the latter is that there is no need to compute an additional index for the sub-grid; however, most elements of the alternative $S2\_count$ array would never be accessed and would hold just the initial value (zero), which is wasteful utilization of memory. Further, the neighbouring sub-grid concept also saves unnecessary computations and will be discussed in Sec.~\ref{subsubsec:K_reduce_to_coo}.

Lastly, the $sum\_R$ array must be filled with the sum of one-step rewards across $N_{r,v}$ realization. Therefore, all threads associated with a given spatial cell (but different realizations) participate in another $atomicAdd()$ operation (Line~\ref{algline:atAdd2}), adding their one-step rewards to the corresponding index $id2$ of the $sum\_R$ array.  A visual representation of this computation is shown in Fig.~\ref{fig:methodology}(C), where threads along the stack of blocks access and write to appropriate elements of the $sum\_R$ array through the $atomicAdd()$ operation named $At\_add\_2()$.

\subsubsection{K\_COMPUTE\_MEAN}
\label{subsubsec:K_comp_mean}
As per Eq.~(\ref{eqn:R(s,a)=E2}), we must compute the expected one-step rewards. In our approach, we estimate this expectation by taking a sample mean of $R(s,a,s')$ over all realizations. The $sum\_R$ array already contains the sum of all $R(s,a,s')$ over all realizations. Hence, the expectation can simply be computed by  dividing each element in $sum\_R$ by $N_{r,v}$. The $K\_COMPUTE\_MEAN$ kernel performs this division operation through a 1D grid of thread-blocks, where each thread-block is also one dimensional.
Each thread is assigned uniquely to an element in $sum\_R$ and simply divides its corresponding element by $N_{r,v}$, thus resulting in 
$R_{a,t}$, which constitutes the expected one-step rewards for performing the action $a$ in each state at time $t$. Fig.~\ref{fig:methodology}(C) and Fig.~\ref{fig:methodology}(D) show the $K\_COMPUTE\_MEAN$ kernel using $sum\_R$ array and modifying it to the resulting $R_{a,t}$ array.

\subsubsection{K\_COUNT}
\label{subsubsec:K_count}


We must now compute the transition probability matrix $P_{a,t}$ for transitions from state $s$ at time $t$ to the states $s'$ at the successive time $t'$, for a given action $a$. As $P_{a,t}$ is sparse, we store it in the sparse matrix COO (coordinate) format that needs only $O(3nnz)$ memory (in contrast to $O({N_c}^2)$ for a full matrix), where $nnz$ is the number of non-zero elements in the sparse matrix. However, we must first compute $nnz$ to allocate the appropriate amount of memory for $P_{a,t}$. We must also separately store the number of non-zero elements in each neighbouring sub-grid of each state $s$ at time $t$ in an array called the $nnz\_count$ array, which is used for accessing the appropriate index of $P_{a,t}$ in the next kernel (Sec.~\ref{subsubsec:K_reduce_to_coo}). The $K\_COUNT$ kernel works as follows. It is launched using a 1D grid of $N_c$ thread-blocks, where each thread-block containing $N_{sg}$ threads. Hence, the total number of threads is $N_c\times{N_{sg}}$, one for each element of the $S2\_count$ array. For example, threads from a given thread-block would be responsible for accessing their respective red highlighted elements in the $S2\_count$ array, as shown in Fig.~\ref{fig:methodology}(D). 
Each thread checks if its corresponding element of the $S2\_count$ array holds a non-zero value. If so, it participates in an atomicAdd() operation, incrementing the count maintained at the appropriate index of the $nnz\_count$ array. Then $nnz$ is directly computed using the Thrust \cite{Thrust} library's parallel implementation of the $reduce()$ function.

\subsubsection{K\_REDUCE\_TO\_COO}
\label{subsubsec:K_reduce_to_coo}


The $S2\_count$ and $nnz\_count$ 
arrays together encode all the information required to populate $P_{a,t}$ in the COO format, which is a matrix of size $3\times{nnz}$.
The first row of $P_{a,t}$ stores all the row indices ($s$ values), the second row stores the column indices ($s'$ values), and the third row stores the non-zero element value (the transition probability $P(s'|s,a)$) .
Fig.~\ref{fig:methodology}(D) shows the transition probability matrix in the COO format, with three $nnz$-sized rows for $s$, $s'$ and $P(s'|s,a)$, respectively.

The $K\_REDUCE\_TO\_COO$ kernel's purpose is to reduce the encoded state transition information in the $S2\_count$ array to the transition matrix $P_{a,t}$. The kernel is launched with $N_c$ threads organised into a 1D grid of thread-blocks, where each thread-block is also one dimensional. Hence, each thread is responsible for computation involving its own initial state $s$.
Each thread first computes its one-dimensional state index $s\_id_{x,y,t}$ and another index $P\_id$ to access the appropriate element of $P_{a,t}$. Then it writes the computed value to $nnz\_s$ successive elements starting from the index $P\_id$.
Here $nnz\_s$ is the number of possible successor states $s'$, for the state-action pair $(s,a)$, and obtained by accessing $nnz\_count$ array. 
Then, to fill the second and third rows of $P_{a,t}$, each thread traverses through the neighbouring sub-grid of its corresponding initial state $s$ in the $S2\_count$ array to check for non-zero values. 
When a thread encounters a non-zero value, it decodes the one-dimensional index of the successor state, $s\_id_{x',y',t'}$
and writes the value to the appropriate index of $P_{a,t}$'s second row. It also maintains a local count of the number of times each $s'$ occurs.
Finally, the transition probability is computed by dividing the count of each $s'$ by the total number realizations $N_{r,v}$ and written at the same index of $P_{a,t}$'s third row.

We briefly mentioned in Sec.~\ref{subsubsec:K_one_step} that the neighbouring sub-grid concept also saves unnecessary computation apart from saving memory. The reason is that in our approach, a thread has to traverse through only $N_{sg}$ elements, whereas, in the naive alternative with no neighbouring sub-grid, a thread would have to traverse through $N_c$ elements, most of which would be $0$. Since the size of the neighbouring sub-grid is much smaller than that of the spatial grid ($N_{sg} << N_c$), the cost of searching is significantly cheaper in our approach compared to the naive alternative.

In Algorithm~\ref{algo:building the model}, the above sequence of kernels is executed within two for-loops. After the last kernel $K\_REDUCE\_TO\_COO$ is executed, we have both $P_{a,t}$ and $R_{a,t}$ stored in the GPU's global memory. At the end of the inner action loop, $P_{a,t}$ and $R_{a,t}$ are moved to the CPU memory through a MEMCPY operation to maintain sufficient memory in the GPU to store similar data structures that will be populated in the next iteration. In addition, $P_{a,t}$s and $R_{a,t}$s of all actions are appended across time-steps in the CPU asynchronously with respect to the GPU. Hence, at the end of the two-level nested for-loops, we have the complete transition matrix $P_a$ and expected one-step rewards $R_a$ for all the states in the spatio-temporal grid for each action $a$. Finally, we append all $P_a$ and $R_a$ across actions along rows and obtain $\underline{P}$ and $\underline{R}$. This final assembly is also done in the CPU and is essentially a pre-processing step for providing a suitable input data structure to the Sparse Value Iteration solver.

\subsection{Sparse Value Iteration on GPUs}
The optimal value function and policy for the above MDP model ($\underline{P}$, $\underline{R}$) is computed by a dynamic programming algorithm called Sparse Value Iteration \cite{sapio2018efficient} that exploits properties of sparse transition matrices. 
We modify \cite{sapio2018efficient}'s GPU implementation algorithm (code available at \cite{spvi.cu_Asapio}) to seamlessly integrate it with our algorithm for phase-1. The modifications include the use of different data structures such as the sparse COO format for storing matrices, change in functions responsible for reading input data and multiple variable names.

Overall, our novel model-building algorithm and sparse value iteration algorithm make an end-to-end GPU software package that takes the stochastic environment's data as input and provides an optimal policy for the MDP path planning problem as output, orders of magnitudes faster than reported by conventional CPU methods. In the following section, we demonstrate applications using our end-to-end software.

\section{Applications}
\label{sec:Applications}
\begin{figure*}
\centering
\includegraphics[width=0.9\textwidth]{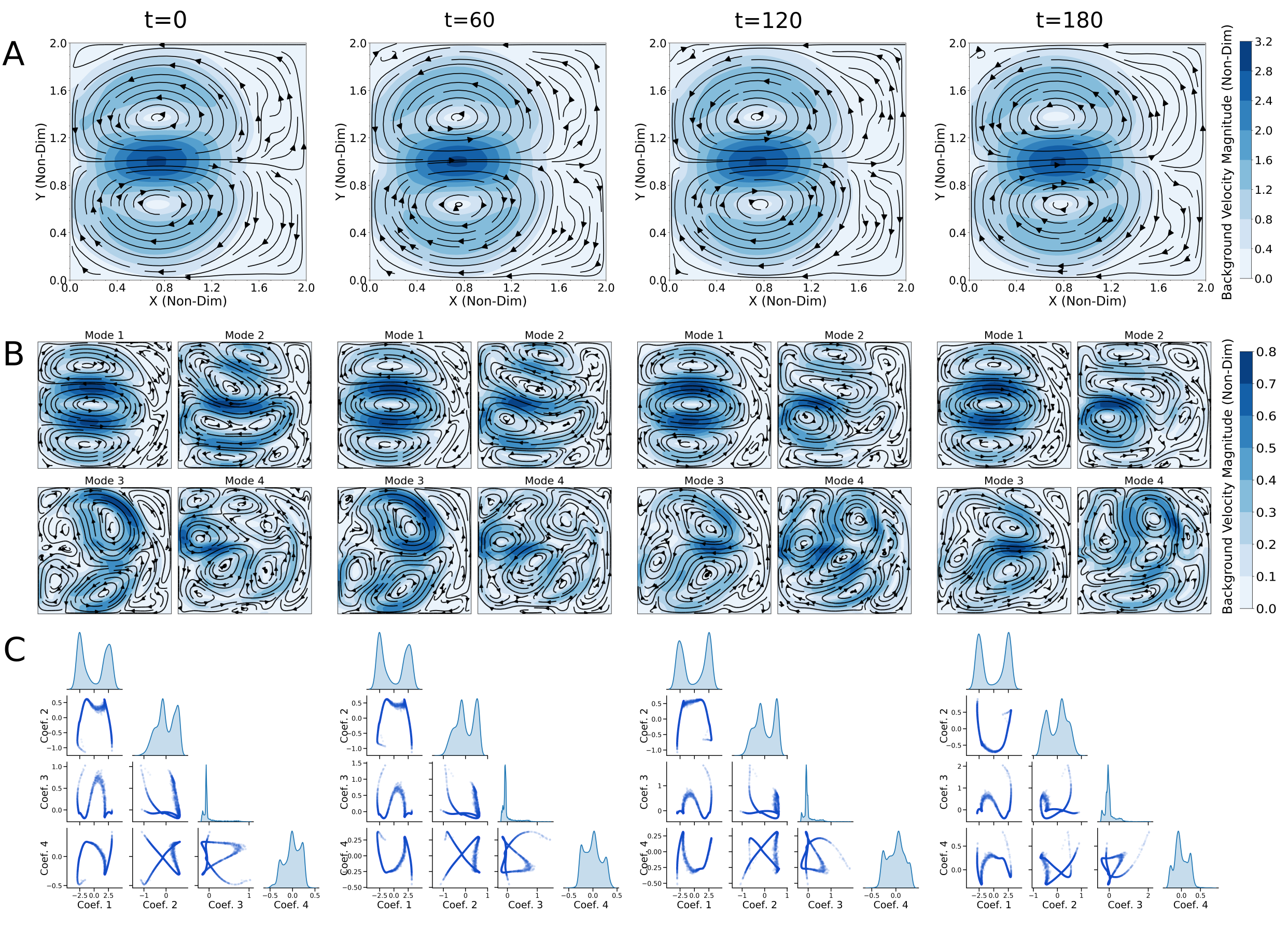}
\caption{ \textbf{Double-gyre stochastic dynamic velocity field.} {(A)} The mean velocity field $\bar{\mathbf{v}}(\mathbf{x},t)$ shown as streamlines overlaid on a background colored with the magnitude. {(B)} Four dominant modes $\tilde{\mathbf{v}}_i(\mathbf{x},t), i=1,2,3,4$ shown as streamlines overlaid on a background colored with the magnitude. {(C)} Pairplots showing the pairwise joint pdf of the DO coefficients $\mu_i(t;\omega_v)$ corresponding to the first four modes. All fields and coefficients shown at four non-dimensional time units. Individual realizations of the velocity fields can be reconstructed from this prediction using the DO relation $\mathbf{v}(\mathbf{x},t;\omega_v)=\bar{\mathbf{v}}(\mathbf{x},t)+\mu_i(t;\omega_v)\tilde{\mathbf{v}}_i(\mathbf{x},t)$, where $\omega_v$ is a sample from the joint PDF of the coefficients.}
\label{fig:dofield_vs_time}
\end{figure*}

\begin{figure*}
\centering
\includegraphics[width=0.9\textwidth]{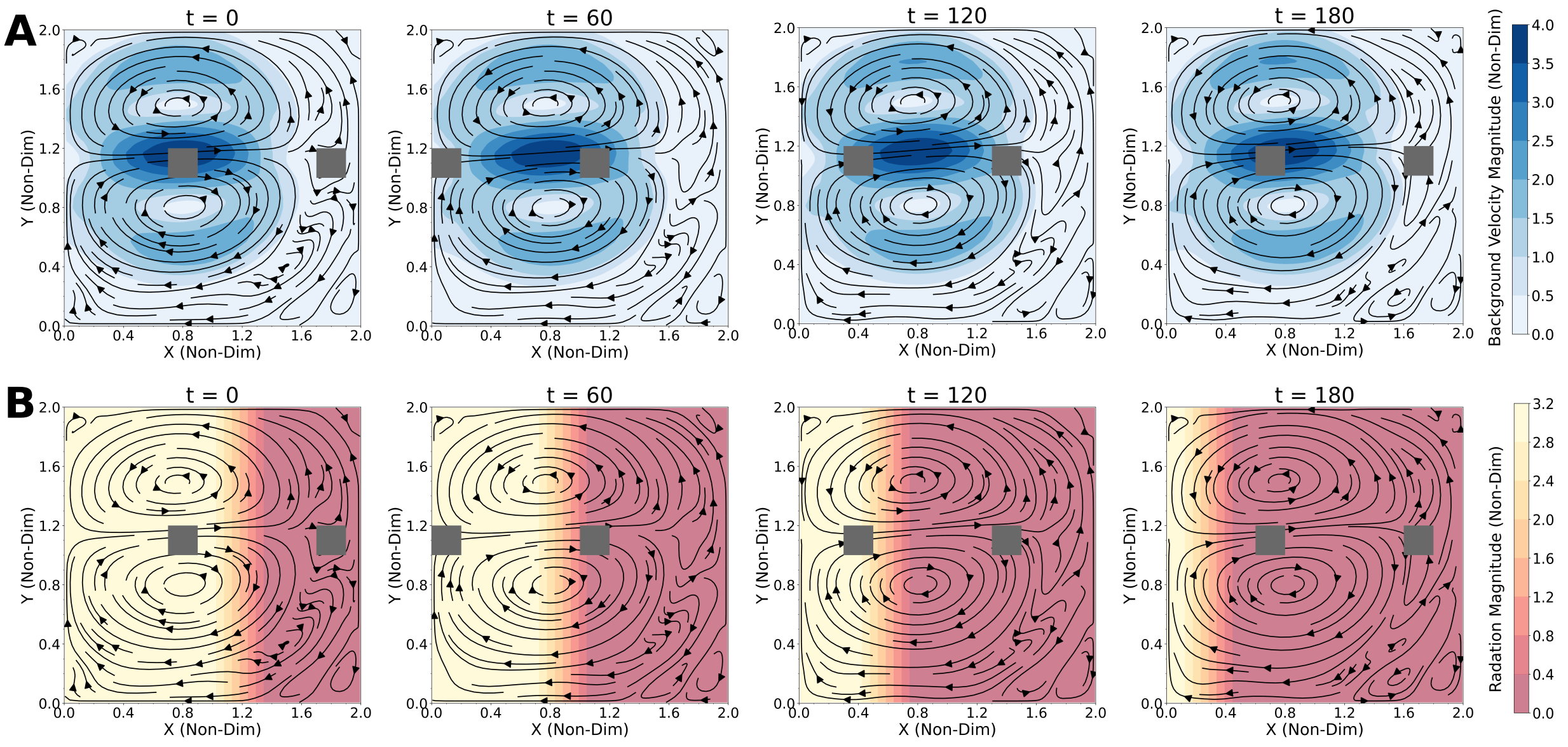}
\caption{ \textbf{Evolution of the environment with time.} (A) Four time snapshots of a particular realization of the stochastic double gyre flow shown as streamlines overlaid on a colorplot of the magnitude. (B) Four time snapshots of the scalar radiation field shown as a colorplot overlaid with the streamlines of the velocity realization. The dynamic obstacle is shown in grey filled patches. }
\label{fig:complete_env}
\end{figure*}

We demonstrate three different single objective missions (Sec.~\ref{sec:MDPFramework}) in a stochastic double-gyre ocean flow simulation. This flow field is obtained by solving the dynamically orthogonal quasi-geostrophic equations \cite{sapsis_lermusiaux_PD2009} and has been extensively used for method development and testing in several path planning \cite{subramani_lermusiaux_CMAME2019} and uncertainty quantification \cite{feppon_lermusiaux_SIREV2018} applications. The equations and description of the physical process associated with the flow field are in \cite{subramani_et_al_CMAME2018}. Briefly, the flow field corresponds to that in a square basin of size 1000 km $\times$ 1000 km, where a wind from west to east sets up a double-gyre flow and the uncertainty arises from its initial conditions. The discrete spatial domain is a square grid of size $200\times 200$, with $\Delta x = \Delta y = 1$ non-dimensional units. We also use 200 discrete time intervals with $\Delta t = 1$ non-dimensional units, thereby making our spatio-temporal grid of size $200\times 200\times 200$. The non-dimensionalization is done such that one non-dimensional unit of space is 5 km and time is 0.072 days. The environment is common across all three applications. Fig.~\ref{fig:dofield_vs_time} shows the mean (row A), the first four DO modes (row B), and the joint-PDF of the first four DO coefficients (row C) at four discrete times (in four columns), highlighting the dynamic and non-Gaussian nature of the stochastic flow field. Fig.~\ref{fig:complete_env} shows the complete environment for a particular realization of the background velocity field  (Fig.~\ref{fig:complete_env}A) and the scalar energy field  (Fig.~\ref{fig:complete_env}B). The stochastic scalar field is assumed to be that of solar radiation incident upon the ocean surface, mimicking a cloud cover moving westwards. This field plays a role only for the net-energy consumption minimization goal. The dynamic obstacles are assumed to be restricted regions in the domain that changes deterministically with time. For the present applications, the dynamic obstacle is square-shaped with a side of 20 units in the $200\times 200$ grid. The obstacles enter the domain from the west and move eastward with time. It is also assumed that these restricted regions do not interact with or affect the flow in any way. In case the movement of obstacles is uncertain and/or the scales of the flow are such that the obstacles affect the flow, these assumptions can be relaxed and environmental flow fields solved with the obstacles as done in \cite{subramani_et_al_CMAME2018}. We consider that the agent can take one of 32 actions ($N_h = 16$ and $N_F = 2$). The number of velocity field realizations used here is $N_{r,v} = 5000$. The size of the neighbouring sub-grid is set $N_{sg} = 21\times21$. 
In dimensional units, the maximum magnitude of the flow field is around 2 m/s and the speed of the agent is 80 cm/s. The planned paths are valid for any AUV that flies at the specified speed. Appropriate operational parameters can be chosen to fix this AUV speed \cite{sherman_2001_spray,edwards_et_al_Oceans2017,lermusiaux_et_al_springer_HOE2016}. As such, the flow is relatively strong, and a typical autonomous marine agent modeled here gets strongly advected by the currents, making it difficult for the agent to move against the flow in most parts of the domain. For the test cases demonstrated here, the agent starts at the position $x_0 = (100, 40)$ (represented by a circular marker), and the target is the position $x_f = (100, 160)$ (represented by a star-shaped marker).  

\subsection{Time-Optimal Planning with Dynamic Obstacles}
\label{subsection:Time_otimal_planning}
\begin{figure*}[h]
\centering
\includegraphics[width=0.9\textwidth]{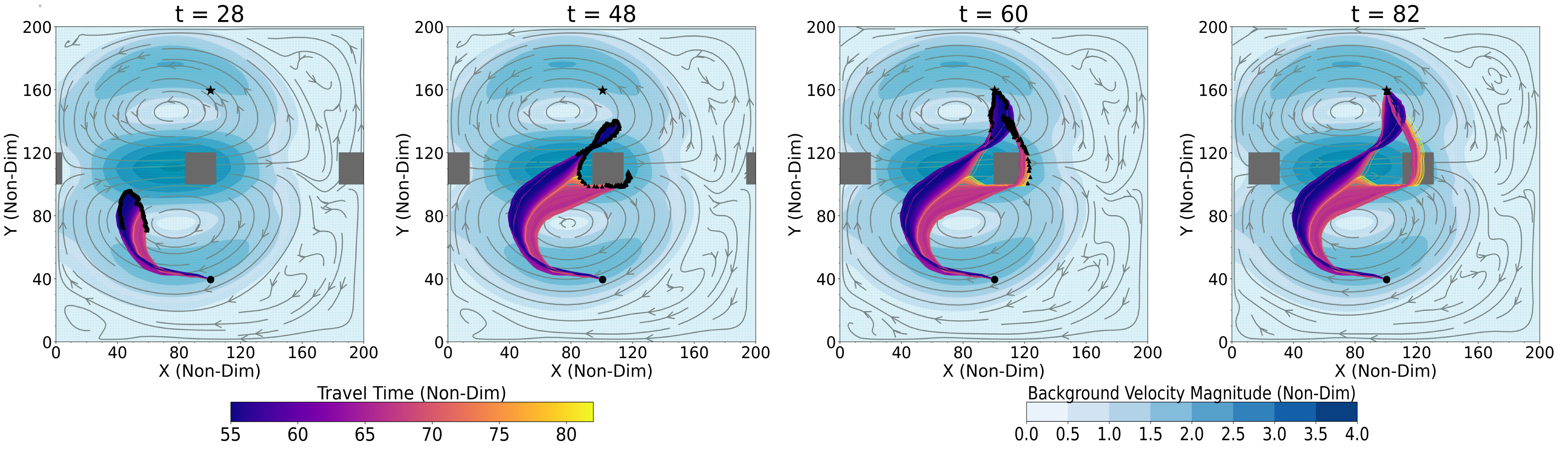}
\caption{\textbf{Paths by following the Time-Optimal Policy}: Distribution of paths obtained by following a time-optimal policy in different realizations of the stochastic environment. The background is the mean fields magnitude with streamlines overlaid to illustrate the nature of flow field encountered. The trajectories are colored with arrival time.}
\label{Fig:Time_Optimal_Trajs}
\end{figure*}

First, we consider a mission with the objective of minimizing the agent's expected travel time from the start to target in the environment described above. We use the one-step rewards defined in Eq.~\ref{eqn:Rsa_time} for time-optimal planning. Our novel end-to-end GPU-implemented algorithm is used to compute the time-optimal policy. This policy is then applied to the ensemble of flow fields, and the computed trajectories are shown in Fig.~\ref{Fig:Time_Optimal_Trajs}. The four discrete time instances shown illustrate how the different trajectories (each corresponding to a different background flow field) evolve with time. To minimize the travel time, the agent takes the high-speed action throughout the mission while intelligently utilizing the double-gyre flow. The distribution of trajectories splits into two bands, successfully avoiding the dynamic obstacles as seen at t=48 and 60 (Fig.~\ref{Fig:Time_Optimal_Trajs}). The computed expected travel time is 61.25 non-dimensional units (4.41 days). The total computational time required for building the MDP model and solving via the Value Iteration algorithm is of the order of a few minutes. In contrast, the MDP-based CPU implementation for a smaller spatio-temporal grid, without a realistic uncertain environmental flow forecast, reported in \cite{kularatne_et_al_2018ICRA} was around 13 hours. Further, the overall paths are similar to the time-optimal trajectories computed from the level set equations in \cite{subramani_et_al_CMAME2018}. However, the key difference is that using the MDP approach, we get an optimal policy and not simply the distribution of optimal waypoint or heading sequence obtained by solving the stochastic level set equations.
 
\subsection{Minimal Energy Expenditure with Time Constraints}
\label{subsection:nergyConsumtion_optimal_planning}

\begin{figure*}[h]
\centering
\includegraphics[width=0.9\textwidth]{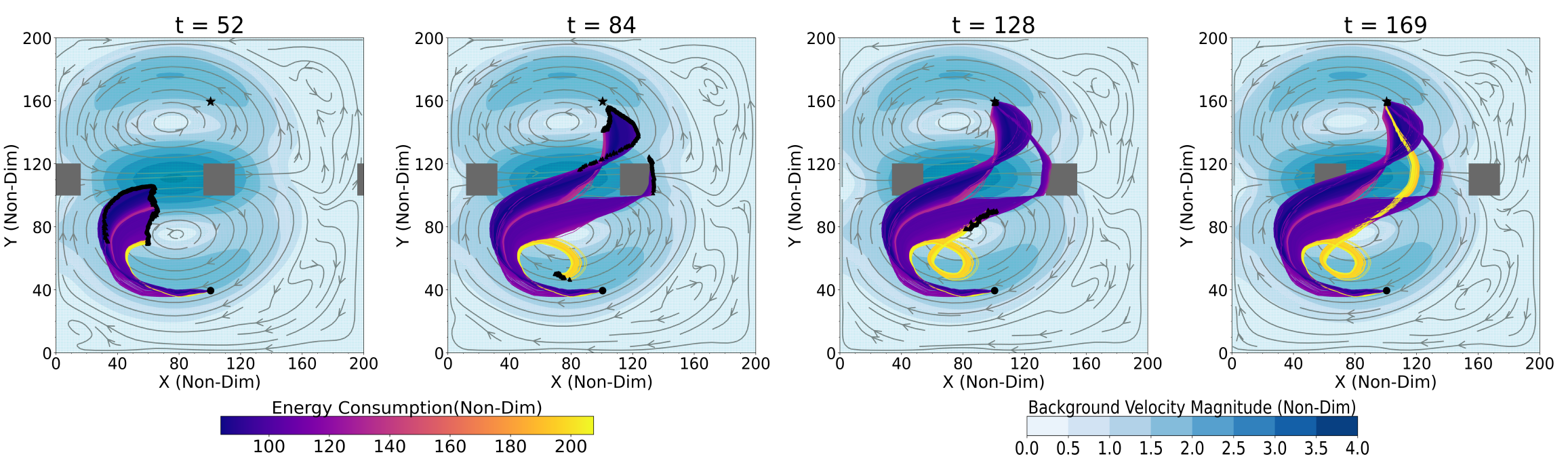}
\caption{\textbf{Paths by following the Energy-Optimal Policy}: Same as Fig.~\ref{Fig:Time_Optimal_Trajs} but for an agent following the energy-optimal policy. Trajectories are colored with the energy consumption.}
\label{Fig:EnergyExpenditure_Optimal_Trajs}
\end{figure*}

Second, we consider a mission with the objective of minimizing the agent's expected energy consumption for travelling from the start to target in the environment described above. An additional constraint is that it must complete its mission within 200 time-steps, which is imposed by limiting the state space to 200 temporal units. Further, the energy consumption model for the agent is assumed to depend only on its speed (Sec.~\ref{subsection:nergyConsumtion_optimal_planning}) and the one-step rewards defined in Eq.~\ref{eqn:Rsa_energyConsumption} is used here. Since we are only optimizing for energy consumption and not energy collection, the scalar field's presence does not affect the optimal policy here. The evolution of the trajectories (at four discrete instances) obtained by following the optimal policy computed by our end-to-end GPU-implemented algorithm is shown in Fig.~\ref{Fig:EnergyExpenditure_Optimal_Trajs}. In contrast with the behaviour observed in Sec.~\ref{subsection:Time_otimal_planning}, the agent here typically takes the slow-speed action throughout the mission to reduce its energy consumption. Due to its low-speed actions, a relatively high background flow, and the presence of two dynamic obstacles, the distribution of trajectories splits into three bands. Two of these bands go around the dynamic obstacle, avoiding it from either side, similar to what we observed in Fig.~\ref{Fig:Time_Optimal_Trajs}. However, the third band (coloured yellow) has to swerve around, making a loop before moving towards the target, since otherwise, it would possibly bump into the dynamic obstacle. The agent's expected energy consumption is 108.7 non-dimensional units. Our framework is flexible to allow using different speeds for the agent based on its capabilities. To do so, we simply have to change the number of discrete speeds $N_F$ considered.

\subsection{Net-Energy-Optimal Planning with Dynamic Intake}
\label{subsection:NetEnergy_optimal}
Third, we consider a mission with the objective of minimizing the agent's expected net energy consumption for travelling from the start to target in the environment described above. The net energy consumption is the difference between the energy consumed by the agent's motors for manoeuvring and the energy collected from solar radiation through its solar panels. In this case, the scalar radiation field plays a significant role in the optimization problem, and the one-step rewards defined in Eq.~\ref{eqn:Rsa_NetEnergy} is used in this case.

\begin{figure*}[h]
\centering
\includegraphics[width=0.9\textwidth]{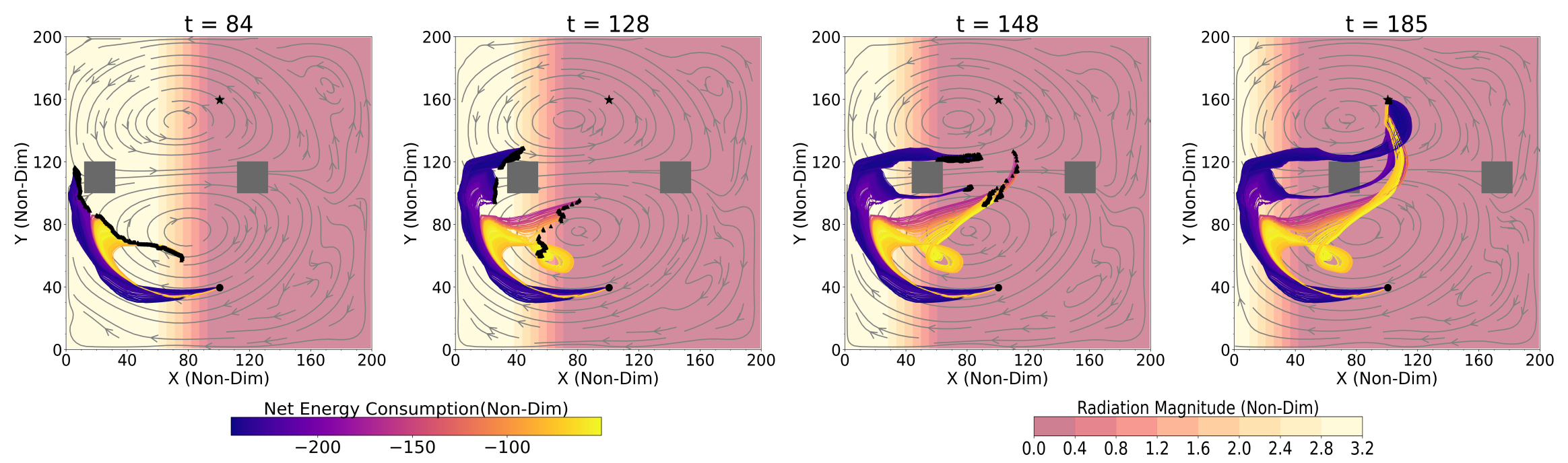}
\caption{\textbf{Paths by following the Net-Energy-Optimal Policy}: Same as Fig.~\ref{Fig:Time_Optimal_Trajs} but for an agent following the net-energy-optimal policy. Trajectories are colored with the net energy consumption.}
\label{Fig:NetEnergy_Optimal_Trajs}
\end{figure*}

Fig.~\ref{Fig:NetEnergy_Optimal_Trajs} shows the trajectories obtained by following the optimal policy for net energy consumption in different realizations of the flow field. The computed optimal policy induces a tendency in the agent to manoeuvre in the yellow sunny region and collect solar energy as long as possible while ensuring that it completes the mission within the given time constraint (Fig.~\ref{Fig:NetEnergy_Optimal_Trajs}). The vehicle starts by moving to the extreme left of the domain to collect solar energy in the yellow sunny region for all realizations of the background environment. The distribution of trajectories initially splits into two bands, as seen at t=48 (Fig.~\ref{Fig:NetEnergy_Optimal_Trajs}). One  band (purple colors) continues to stay in the left extreme of the domain. The other  band (yellow colors) moves back towards the right and then loops around to collect energy in the sunny region as much as possible. At around $t=128$, the diminishing sunny region and the relatively strong background flow make it infeasible for the vehicle to keep collecting solar energy on the left side of the domain. After that, the vehicle starts moving towards the target location. On the way, the vehicle in the realizations corresponding to the purple band encounters the dynamic obstacle. In response, the band further splits into two, moving past above and below it. Finally, all trajectories move towards the target location and converge there by $t=185$. The average net energy consumption is -190 non-dimensional units, indicating that more energy is harvested than expended for the particular non-dimensional constants used here.

Crucially, all these computations were performed on a modest GPU in a fraction of the time required by conventional MDP model builders and solvers using a CPU. A detailed computational run-time study is presented next. 
\subsection{Run-Time Analysis with Different GPUs}
The above applications were run for various problem sizes on three different GPUs, namely RTX 2080, RTX 8000 and a V100. The execution times were compared with the sequential version of the algorithm run on an Intel Core i7-8700 3.2 0GHz CPU with 32 GB RAM. The GPU runs were individually optimized for launching kernels at the optimal threads per block. These were found to be 64, 64 and 32 for the RTX 2080, RTX 8000 and V100, respectively. Our model-building algorithm's execution times and speedups were noted for a varying number of states, number of actions, and number of realizations of the stochastic velocity field. 

\begin{figure*}[ht]
\centering
\includegraphics[width=0.9\textwidth]{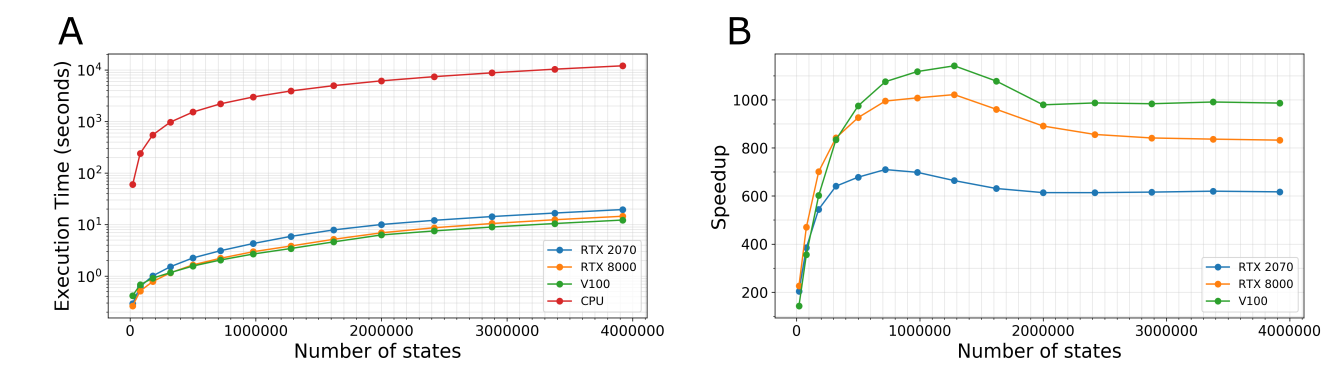}
\caption{\textbf{Run-time analysis of the algorithm}.\textbf{(A)} Semi-log plot of execution time vs number of states for different GPUs. \textbf{(B)} Plot of speedup vs number of states for different GPUs}
\label{fig:ExecTime_Speedup_vs_Nstates}
\end{figure*}

Fig.~\ref{fig:ExecTime_Speedup_vs_Nstates}(A) shows the execution time involved in model building for different problem sizes (number of states) for all the GPU and CPU runs. Fig.~\ref{fig:ExecTime_Speedup_vs_Nstates}(B) shows the speedup obtained by the GPU when compared to the CPU. We see that our parallel model-building algorithm's execution time over the different GPUs is 2-3 orders of magnitude faster than the sequential version. Correspondingly, for a large number of states, e.g., 10 million, we see that we get speedups of approximately 600, 820 and 1000 on the RTX 2080, RTX 8000 and a V100, respectively. As expected, similar speedups were observed when execution times were compared by varying the number of actions and the number of realizations (not shown here) used to characterise the stochastic velocity field.

The speedup depends on the GPU specifications such as the number of available CUDA cores, GPU clock rate, GPU architecture, and other technological advancements. It also depends on the relative amount of serial work each thread has to perform. In our case, loosely speaking, the amount of serial computation per thread is relatively small. Hence, the V100 outperforms the other GPUs as it has a larger number of CUDA cores, despite having a lower clock rate.

The trend shows that even for large problem sizes, we can build MDP models in the order of a few minutes. The computation times are smaller than what we reported in \cite{chowdhury2020DDDAS}, where we used a CPU implementation of a model-based reinforcement learning algorithm we devised. This is promising as it enables us to frequently update the MDP model in real-time for common marine applications. Updating the MDP model would require in-mission measurements by the agent from the environment and an efficient implementation of a data assimilation scheme to compute the posterior environmental forecast on the go. The posterior environmental forecast can then be used to update the MDP model and the optimal policy in-mission using our fast and scalable GPU algorithm.

\section{Conclusion} \label{sec:conclusion}
We developed an end-to-end GPU algorithm for optimal path planning of autonomous agents in a strong, stochastic, dynamic flow field. Different objective functions for the optimal planning viz., time, energy consumed and net energy consumed in the presence of an energy field were considered. Our primary contribution in the present paper was the development of a GPU implemented fast and scalable solver for building the MDP model state transition probability and rewards. Another contribution is that we consider a net energy consumption objective under an uncertain scalar energy field. We designed and implemented an algorithm that comprises multiple kernels for computing the transition probability and one-step rewards of the agent. We prove that for the scalar field only the mean of the environment is required in one step reward calculation. We also introduced the neighbouring sub-grid concept that helps save memory and unnecessary computation and plays a key role in making our kernels computationally efficient.


To illustrate our planner's working, we showcase the optimal policy computed in a stochastic double-gyre flow field for thee single-objective missions. In the future, our planning framework  can be utilized for multi-objective planning where all competing objectives are optimized together to form Pareto optimal fronts. Our planner can also be used for pursuit-evasion games \cite{sun_et_al_JGCD2017}, ship routing \cite{mannarini_et_al_2016visir}, drone delivery systems and traffic management \cite{ong2015short,koster2018anticipating}. A real-time sea exercise to test our new planner can also be undertaken, similar to the field experiments for our deterministic level-set planner \cite{subramani_et_al_Oceans2017}. Most importantly, we plan to extend the framework to perform on-board routing where the MDP model is updated with new observations of the environment, and an updated policy is re-planned quickly using our efficient GPU algorithm on-board the vehicle during a mission. Further, if any GPU-based RL algorithms are developed \cite{babaeizadeh2016reinforcement}, our solver and results can be used for benchmark of the approximate RL solution.
\bibliographystyle{IEEEtran}
\bibliography{refs,mybibfile,mseas}



\section*{Acknowledgment}
The authors would like to thank the Prime Minister's Research Fellowship (PMRF) for supporting the graduate studies of RC. We also acknowledge the partial support received for the present work from INSPIRE Faculty Award, Arcot Ramachandran Young Investigator Award and an internal grant of IISc.

\ifCLASSOPTIONcaptionsoff
  \newpage
\fi

\begin{IEEEbiographynophoto}{Rohit Chowdhury} received his B.Tech degree in Mechanical Engineering from NIT Warangal, and is currently a Ph.D. candidate at the Dept. of Computational and Data Sciences, IISc. His research interests include parallel programming, optimal path planning and data assimilation. He is a recipient of the prestigious Prime Minister's Research Fellowship, Government of India.
\end{IEEEbiographynophoto}

\begin{IEEEbiographynophoto}{Deepak Subramani} received his Ph.D. in Mechanical Engineering and Computation from the Massachusetts Institute of Technology, Cambridge, MA, USA. His research interests are in the use of Machine Learning and Artificial Intelligence in Geosciences, data assimilation, autonomous routing of marine vehicles and data-driven application systems. He is a recipient of several awards including the de Florez research award, SNAME outstanding research award and Arcot Ramachandran Young Investigator Award for his work on optimal path planning of marine vehicles.
\end{IEEEbiographynophoto}





\end{document}